\documentclass{vgtc}                          % final (conference style)
\ifpdf%                                % if we use pdflatex
  \pdfoutput=1\relax                   % create PDFs from pdfLaTeX
  \usepackage{graphicx}                % allow us to embed graphics files
  \DeclareGraphicsExtensions{.pdf,.png,.jpg,.jpeg} % for pdflatex we expect .pdf, .png, or .jpg files
\else%                                 % else we use pure latex
  \usepackage{graphicx}                % allow us to embed graphics files
  \DeclareGraphicsExtensions{.eps}     % for pure latex we expect eps files
\fi%

\graphicspath{{figures/}{pictures/}{images/}{./}} % where to search for the images

\usepackage{microtype}                 % use micro-typography (slightly more compact, better to read)
\PassOptionsToPackage{warn}{textcomp}  % to address font issues with \textrightarrow
\usepackage{textcomp}                  % use better special symbols
\usepackage{mathptmx}                  % use matching math font
\usepackage{times}                     % we use Times as the main font
\usepackage{cite}                      % needed to automatically sort the references
\usepackage{tabu}                      % only used for the table example
\usepackage{booktabs}                  % only used for the table example
\usepackage{bm}
\usepackage{amsfonts}
\usepackage{amsmath}
\usepackage{comment}

\onlineid{1056}

\vgtccategory{IEEE VIS Short Paper}

\vgtcinsertpkg

\DeclareMathOperator*{\mlp}{MLP}
\DeclareMathOperator*{\pe}{PE}

\newenvironment{myitemize}{
\begin{itemize}
 \setlength{\itemsep}{1pt}
 \setlength{\parskip}{0pt}
 \setlength{\parsep}{0pt}}{\end{itemize}}
 
\title{FCNR: Fast Compressive Neural Representation of Visualization Images}

\author{Yunfei Lu\thanks{e-mail: ylu25@nd.edu} %
\and Pengfei Gu\thanks{e-mail: pgu@nd.edu} %
\and Chaoli Wang\thanks{e-mail: chaoli.wang@nd.edu}}
\affiliation{\scriptsize University of Notre Dame}

\abstract{We present FCNR, a fast compressive neural representation for tens of thousands of visualization images under varying viewpoints and timesteps. The existing NeRVI solution, albeit enjoying a high compression ratio, incurs slow speeds in encoding and decoding. Built on the recent advances in stereo image compression, FCNR assimilates stereo context modules and joint context transfer modules to compress image pairs. Our solution significantly improves encoding and decoding speed while maintaining high reconstruction quality and satisfying compression ratio. To demonstrate its effectiveness, we compare FCNR with state-of-the-art neural compression methods, including E-NeRV, HNeRV, NeRVI, and ECSIC. The source code can be found at \url{https://github.com/YunfeiLu0112/FCNR}.} % end of abstract

\begin{document}


\vspace{-0.1in}
\firstsection{Introduction}

\maketitle

%% \section{Introduction} %for journal use above \firstsection{..} instead
%In scientific disciplines, 
Generating vast datasets has become indispensable for analyzing complex phenomena across diverse fields. As technical advances continue to shine in an era of unprecedented data acquisition, managing and interpreting such enormous amounts of data becomes increasingly challenging. Scientific visualization is important for us to visually comprehend complicated patterns, identify trends, and extract meaningful insights from simulation data, which are often time-varying. 
%However, the sheer data size often presents difficulties in storage, transmission, and computation.
% 
When dealing with a time-varying volumetric dataset, we can produce numerous visualization images, including those generated through isosurface rendering (IR) or direct volume rendering (DVR). These images correspond to various parameters, such as 
%isovalues, transfer functions, 
timesteps and camera views, 
offering a thorough representation of the data. 
We tackle the issue of managing significant quantities of these visualization images, which occupy substantial gigabytes of storage and could surpass the original data size. 
This scenario poses significant constraints, including storage cost, network transmission, image access, and interactive display when conveying the visualization output. 
Hence, efficient compression and sharing of these visualization images become a necessity.
Recent developments in DL4SciVis~\cite{Wang-TVCG23} provide a viable direction. 

To achieve this goal, we present FCNR, a \underline{\bf f}ast \underline{\bf c}ompressive \underline{\bf n}eural \underline{\bf r}epresentation for tens of thousands of visualization images. 
FCNR takes a pair of visualization images with nearby views as input. 
It encodes the image pair into quantized bitstreams with entropy coding computed and their similarity exploited. 
It then decodes them to reconstruct the images. 
FCNR can efficiently compress a vast array of images derived from time-varying datasets under different viewing %(i.e., spherical angle coordinates $\psi$ and $\phi$ for sampling viewpoints), 
and timestep
%, and isovalue (for IR) or transfer function (for DVR)
parameters in a relatively short time. 
We evaluate FCNR on multiple datasets, quantitatively and qualitatively, and compare it with state-of-the-art deep learning compression baselines, including E-NeRV, HNeRV, NeRVI, and ECSIC, to demonstrate its superiority. Our contributions are as follows:
\begin{myitemize}
\vspace{-0.1in}
    \item simultaneously compressing a pair of images with similar views based on {\em joint context transfer modules} (JCTMs), which extract mutual information from the whole images; % instead of in the epipolar direction between both images.
    \item incorporating viewpoints and timesteps into {\em stereo context modules} (SCMs) to accommodate our compression scenario while improving entropy estimation of one encoded image using another as the context; 
    \item improving encoding and decoding speeds significantly while maintaining high reconstruction quality and satisfying compression ratio expressed in {\em bit per pixel} (BPP); 
    \item leveraging the interpolation ability by compressing all the images of the given dataset after training on its subset, which also expedites the encoding process.
\vspace{-0.1in}
\end{myitemize}

\vspace{-0.075in}
\section{Related Work}

In recent years, implicit neural representation (INR) has been extensively studied for image and video compression~\cite{chen2021nerv, li2022enerv, chen2022cnerv, zhao2023dnerv, chen2023hnerv, maiya2023nirvana, kwan2024hinerv}. 
%Examples include NeRV~\cite{chen2021nerv}, E-NeRV~\cite{li2022enerv}, CNeRV~\cite{chen2022cnerv}, D-NeRV~\cite{zhao2023dnerv}, HNeRV~\cite{chen2023hnerv}, NIRVANA~\cite{maiya2023nirvana}, and HiNeRV~\cite{kwan2024hinerv}. 
%
NeRV~\cite{chen2021nerv} takes as input an image index, generates the image embedding via multilayer perceptrons (MLPs) and convolutional layers, and outputs the whole image. 
% As an {\em image-wise} method, NeRV significantly improves training and inference speed compared to {\em pixel-wise} methods, especially for videos comprising thousands of high-resolution images. 
%
E-NeRV~\cite{li2022enerv} improves the NeRV architecture by identifying the redundant parts and decomposing the image-wise INR into distinct spatial and temporal contexts, accelerating convergence while maintaining high performance. 
CNeRV~\cite{chen2022cnerv} enables internal generalization using content-adaptive embedding, which compactly encodes visual information. 
D-NeRV~\cite{zhao2023dnerv} represents various videos using the same model, which takes sampled key-frames as input for clip-specific content encoding and outputs video frames with a motion-aware decoder.
%Once trained, each video clip can be reconstructed by feeding time indices and clip-specific key-frames. 
%
HNeRV~\cite{chen2023hnerv} resolves the content-agnostic issue and the unbalanced parameter distribution of NeRV by storing videos in small, content-adaptive frame embeddings and utilizing a learned decoder. % and introducing the novel HNeRV block. 
NIRVANA~\cite{maiya2023nirvana} proposes patch-wise prediction to accommodate videos with varying spatial and temporal resolutions using the same architecture. 
%and achieves less encoding and decoding time while maintaining better or on-par quality at comparable bit per pixel (BPP). 
%
HiNeRV~\cite{kwan2024hinerv} addresses the limited representation capability of INR and refines the model compression pipeline with adaptive parameter weighting and quantization-aware training. 

In scientific visualization, INR has been applied to data generation and compression tasks~\cite{Han-TVCG23,Tang-CG24,Tang-PVIS24}. 
Gu et al.\ \cite{Gu-CG23} extended INR to visualization image compression, which is much more challenging due to the need for accommodating viewpoint and timestep parameters and more significant differences between neighboring rendering images than video frames. The proposed NeRVI achieves neural representations with a high compression ratio and leads to good image fidelity using mask loss. However, the rather slow encoding speed restricts its use in practice, especially when compressing high-resolution visualization images in a large collection. 
%
%Even though INR-based methods compress images with high quality and \hot{compression ratio}, they do not consider the mutual information between images under different views and are costly to train on large datasets, putting them at a disadvantage in visualization image compression. 

\begin{figure*}[htb]
%\vspace{-.1in} 
  \begin{center}
    \includegraphics[width=1.0\linewidth]{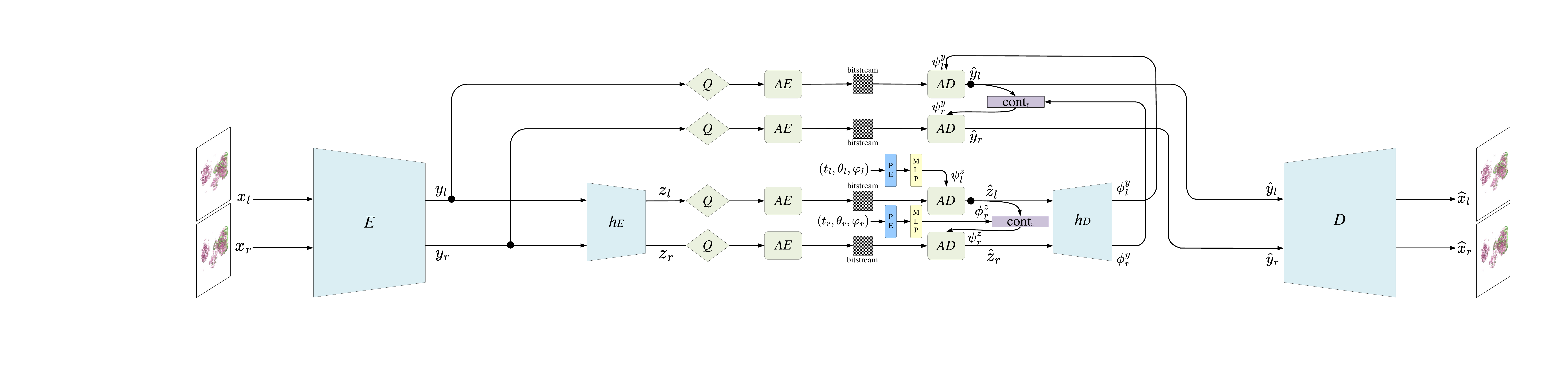}
 \end{center}
\vspace{-.25in} 
 \caption{Overview of FCNR. The encoder ($E$) encodes $x_l$ and $x_r$ to bitstreams with hyper-encoder ($h_E$), quantization ($Q$), and arithmetic encoder ($AE$). The decoder ($D$) then reconstructs $\hat{x}_l$ and $\hat{x}_r$ through quantized latents $(\hat{y}_l$ and $\hat{y}_r)$ with arithmetic decoder ($AD$) and hyper-decoder ($h_D$).} %associated with visualization parameters $(t_l,\theta_l,\varphi_l)$ and $(t_r,\theta_r,\varphi_r)$
 \label{fig:overview}
\end{figure*}
%\vspace{-.1in} 

\begin{figure*}[htb]
%\vspace{-.1in}
  \begin{center}
    $\begin{array}{c@{\hspace{0.2in}}c@{\hspace{0.2in}}c}
    \includegraphics[height=0.65in]{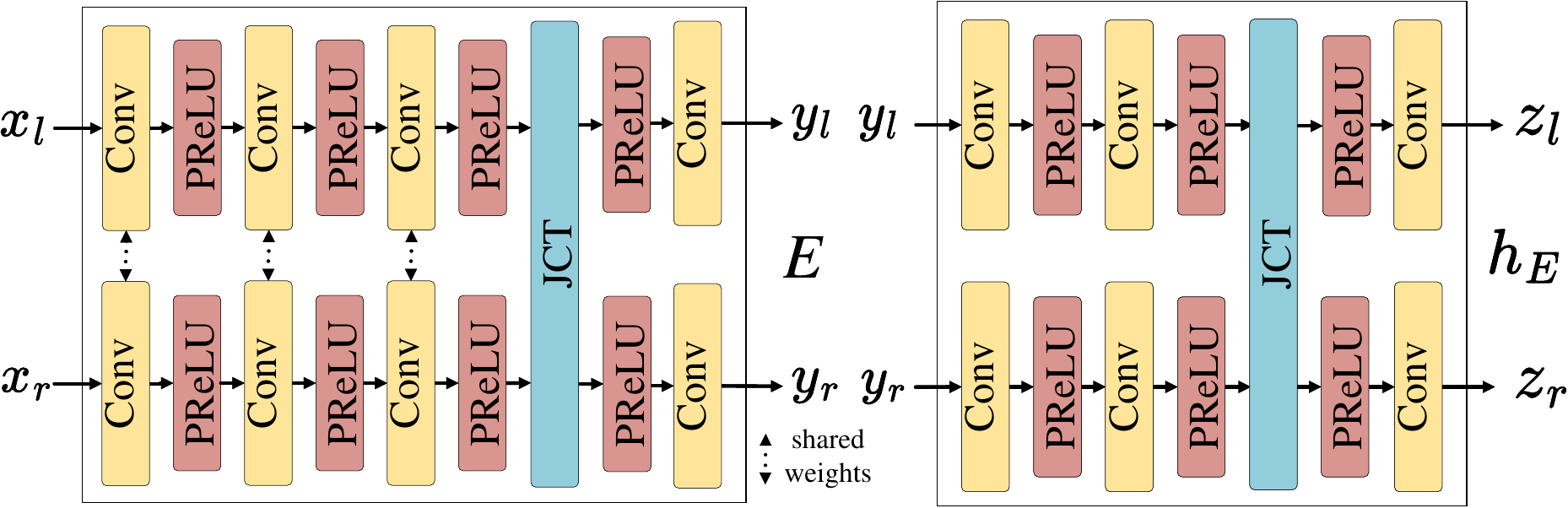}&
    \includegraphics[height=0.65in]{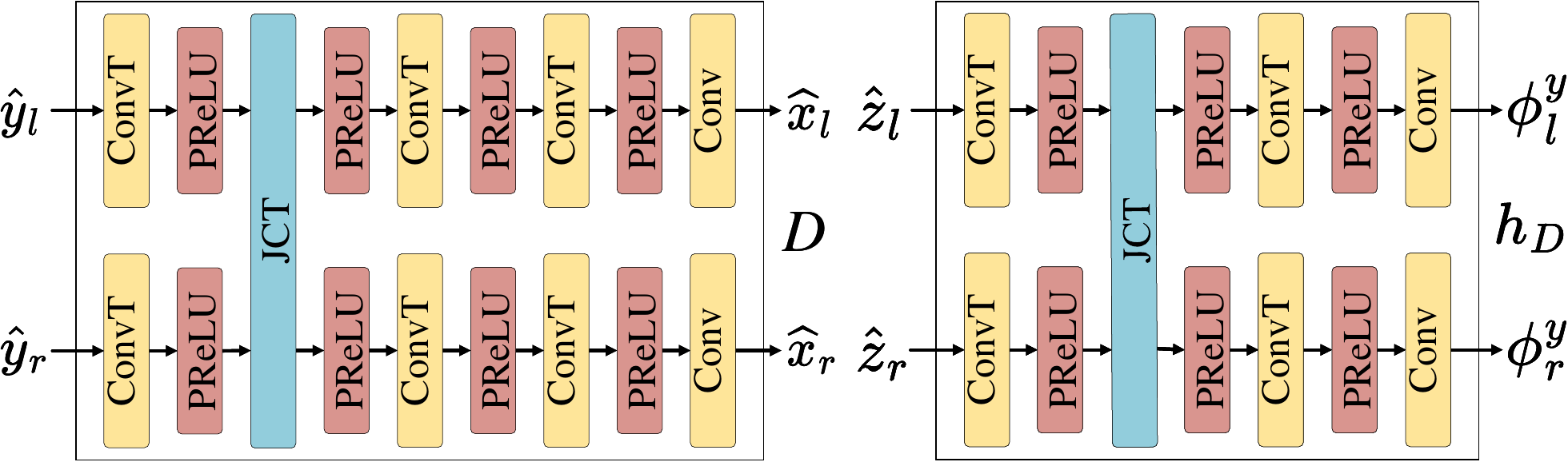}&
    \includegraphics[height=0.65in]{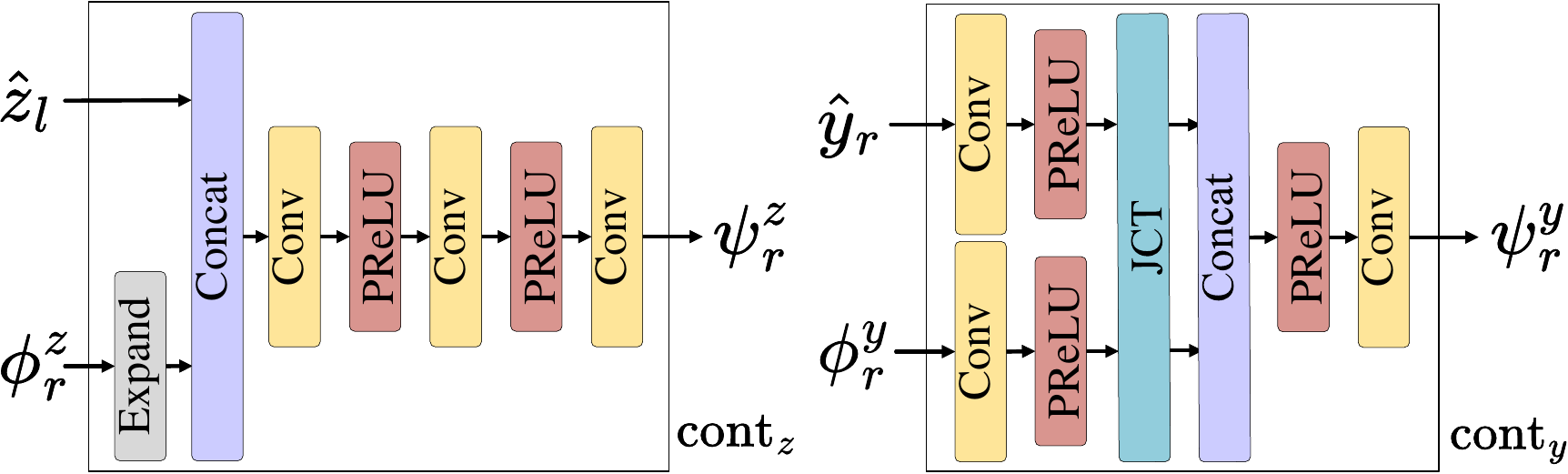}\\
    \mbox{(a) encoder ($E$) and hyper-encoder ($h_E$)}&
    \mbox{(b) decoder ($D$) and hyper-decoder ($h_D$)}&
    \mbox{(c) SCM (cont$_z$) and SCM (cont$_y$)}\\    
    \end{array}$
 \end{center}
\vspace{-.25in} 
 \caption{The detailed structure of each module.} 
 \label{fig:encoders-decoders-context}
\end{figure*}
%\vspace{-.1in} 

In contrast, stereo-image compression methods seek to compress image pairs simultaneously by exploiting their similarities (i.e., mutual information) using neural networks. 
For instance, 
Liu et al.\ \cite{liu2019dsic} presented DSIC, which computes a dense warp field and feeds features from the left image after warping into the encoder and decoder of the right image. 
Deng et al.\ \cite{deng2021deep} designed HESIC, which improves DSIC by applying a rigid image-space homography transform. 
W{\"o}dlinger et al.\ \cite{wodlinger2022sasic} introduced SASIC that enhances a conventional single-image compression backbone model. To accommodate finer local displacements between images, it incorporates latent-domain global shift and subtraction as well as stereo attention modules in the decoder. 
Based on SASIC, W{\"o}dlinger et al.\ \cite{wodlinger2024ecsic} further developed ECSIC that augments the architecture with {\em stereo cross-attention modules} (SCAMs) and SCMs. 
Zhang et al.\ \cite{zhang2023ldmic} proposed LDMIC, a simple and effective cross-attention-based JCTM utilizing the decoder's cross-attention mechanism to capture global inter-view correlations efficiently. 
%
%All these stereo-image compression methods enable the neural network to compress image pairs jointly by exploiting their mutual information. 
In this work, we assimilate the notion of jointly compressing image pairs by exploiting their mutual information. 
We design our network based on ECSIC, incorporate the JCTM from LDMIC for the more complicated visualization image compression task, and present FCNR, a fast solution for compressive neural representation. 

\vspace{-0.075in}
\section{FCNR}

Given a time-varying dataset $Y=\{Y_1, Y_2, \ldots, Y_T\}$, where $T$ is the number of timesteps, with a predefined isovalue or transfer function, we produce a set of visualization images. Each volume $Y_t$, $t \in [1,T]$, is represented as a subset of images $X_t$ associated with different camera views. We aim to learn a mapping that encodes the input images into latent representations, which are quantized, compressed, and decoded to reconstruct the images. 

Networks for stereo image compression often consist of the main autoencoder and hyperprior autoencoder. 
We adapt this structure to compress IR or DVR images under different views and timesteps. 
As shown in Figure~\ref{fig:overview}, the input to our model is a pair of images $x_l$ and $x_r$ ($l$ and $r$ denote left and right) from $X_t$, with the two neighboring views $(\theta, \varphi_l)$ and $(\theta, \varphi_r)$. 
We first use the encoder $E$ to encode $x_l$ and $x_r$ into the latents $y_l$ and $y_r$. Then, we estimate the latent entropy parameters $\psi^y_l$ and $\psi^y_r$ using the hyper-encoder $h_E$ and hyper-decoder $h_D$ to produce quantized latents $\hat{y}_l$ and $\hat{y}_r$. Finally, we utilize the decoder $D$ to reconstruct the images $\hat{x}_l$ and $\hat{x}_r$ from $\hat{y}_l$ and $\hat{y}_r$. 
We improve the structure with SCMs~\cite{wodlinger2024ecsic} (which aid in the prediction of $\psi^y_r$ from $\hat{y}_l$ and hyper-latent entropy parameters $\psi^z_r$ from hyper-latent $\hat{z}_l$) and JCTMs~\cite{zhang2023ldmic} (which exploits the feature-space inter-view correlations brought by overlapping viewpoints in visualization images for generating more informative representations).

{\bf Encoding modules and quantization.}
We develop our $E$ and $h_E$ based on the structure proposed in ECSIC. Each consists of several convolutional (Conv) layers, a JCTM, and parametric ReLU (PReLU) activation functions~\cite{he2015prelu}. Their detailed structures are shown in Figure~\ref{fig:encoders-decoders-context} (a). Unlike the epipolar assumption in ECSIC, the transformations between two visualization images in a pair are much more complex, involving the 3D rotation of the volume. Since SCAM~\cite{wodlinger2024ecsic} only computes cross-attention between the corresponding epipolar lines, it fails to fully capture the inter-view information brought by the more complicated transformations. Therefore, we replace SCAMs with JCTMs to exploit mutual information globally. 
%
%The detailed structure of JCT is shown in Fig $...$, composed of several residual blocks, a Multi-head Cross Attention Module, and skip connections. Each residual block contains two consecutive convolutional layers and leaky ReLU activation functions.
%
Given $x_l$ and $x_r$, $E$ computes the main latent representations $y_l$ and $y_r$ by 
\begin{equation}
y_i=E(x_i), \ i \in \{l, r\}.
\end{equation}
$h_E$ accepts $y_l$ and $y_r$ and generates the hyper-latents $z_l$ and $z_r$
\begin{equation}
z_i=h_E(x_i), \ i \in \{l,  r\}.
\end{equation}

Introducing $z$ for entropy estimation is effective to model the dependencies between $y$, which are assumed to be independently conditioned on $z$~\cite{balle2018variational}. 
$y_l$, $y_r$, $z_l$, and $z_r$ then go through a quantization process. For example, 
\begin{equation}
\hat{y}_i={\rm round}(y_i-\mu^{y}_i)+\mu^{y}_i, \ i \in \{l,  r\},
\end{equation}
where $\mu^{y}_i$ is the estimated mean of the disrtibution of $y_i$. 
A similar process is applied to $z_l$ and $z_r$ to generate quantized hyper-latents $\hat{z}_l$ and $\hat{z}_r$. 
To make the process differentiable, we use approximate quantization by adding the uniform noise $\epsilon \sim \mathcal{U}(-0.5, 0.5)$ for the rate loss~\cite{balle2017end}. 
\begin{equation}
\tilde{y}_i=y_i+\epsilon, \ i \in \{l,  r\}.
\end{equation}
Thus, the density function of $\hat{y}_i$ is a continuous relaxation of the probability mass function of $y_i$, allowing the differential entropy of $\hat{y}_i$ to approximate the entropy of $\hat{y}$. Additionally, independent uniform noise approximates the quantization error, modeling its marginal moments for distortion measurement. $z_l$ and $z_r$ follow a similar process. For the distortion loss, we employ a straight-through-estimation quantization~\cite{minnen2020channel}. 

{\bf Decoding modules and entropy model.}
As shown in Figure~\ref{fig:encoders-decoders-context} (b), $D$ and $h_D$ each have Conv layers, a JCTM, PReLU activation functions, and transposed convolutional (ConvT) layers for upsampling. $\hat{z}_l$ and $\hat{z}_r$ are stored as side information to help predict $\psi^y_l$ and $\psi^y_r$ of $\hat{y}_l$ and $\hat{y}_r$. The distribution of each latent representation is modeled by a Laplacian distribution with parameters $\psi=(\mu, b)$. The process of distribution modeling and parameter estimation corresponds to the arithmetic encoders (AEs) and arithmetic decoders (ADs) shown in Figure~\ref{fig:overview}.
We model the distribution of $\hat{z}_l$ by a channel-wise Laplacian distribution with parameters $\psi^z_l$ computed from visualization parameters $(t_l,\theta_l,\varphi_l)$ 
\begin{equation}
\psi^z_l=\mlp(\pe(t_l,\theta_l,\varphi_l)),
\end{equation}
where PE denotes positional encoding which projects $(t_l,\theta_l,\varphi_l)$ into a higher-dimensional space
\begin{equation}
\pe(u)=(\sin(b^0\pi u), \cos(b^0\pi u),\dots,\sin(b^{L-1}\pi u),\cos(b^{L-1}\pi u)),
\end{equation}
\begin{equation}
\pe(t,\theta,\varphi)=(\pe(t),\pe(\theta),\pe(\varphi)).
\end{equation}
Here, we set $b=1.25$ and $L=8$. 
Since our visualization image pairs have greater variations than the stereo image pairs compressed by ECSIC, computing distribution parameters from visualization parameters can help mitigate the gap by allowing for detailed quantitative differences between images and improve model performance. 
The distributions of $\hat{z}_r$, $\hat{y}_l$, and $\hat{y}_r$ are modeled by factorized Laplacian distributions. 
We further reduce the bitrate by conditioning the distributions of $\hat{y}_r$ and $\hat{z}_r$ on the information from $\hat{y}_l$ and $\hat{z}_l$ with SCMs ${\rm cont}_y$ and ${\rm cont}_z$, and their structures are shown in Figure~\ref{fig:encoders-decoders-context} (c). In this way, $\psi^z_r$ are predicted from $\hat{z}_l$ and parameters $\phi^z_r$
\begin{equation}
\psi^z_r={\rm cont}_z(\hat{z}_l, \phi^z_r),
\end{equation}
where, likewise, $\phi^z_r$ is generated with $(t_r,\theta_r,\varphi_r)$
\begin{equation}
\phi^z_r=\mlp(\pe(t_r,\theta_r,\varphi_r)),
\end{equation}
Then $h_D$ computes $\phi^y_l$ and $\phi^y_r$ from quantized hyper-latents
\begin{equation}
\phi^y_i=h_D(\hat{z}_i), \ i \in \{l, r\}.
\end{equation}
In the main branch, we set $\psi^y_l=\phi^y_l$. For $\psi^y_r$, we similarly apply the SCM
\begin{equation}
\psi^y_r={\rm cont}_y(\hat{y}_l, \phi^y_r).
\end{equation}
Finally, $D$ reconstructs $\hat{x}_l$ and $\hat{x}_r$ from $\hat{y}_l$ and $\hat{y}_r$
\begin{equation}
\hat{x}_i=D(\hat{y}_i), \ i \in \{l, r\}.
\end{equation}

%===================================
\begin{table}[b]
\caption{The resolution and total sampled images of each dataset. ``\# st" denotes the number of timesteps we subsample from the dataset.}
\vspace{-0.05in}
\centering
%{\scriptsize
\resizebox{\linewidth}{!}{
\begin{tabular}{ccccc}
dataset & resolution ($x \times y \times z \times t$)  & \# views & \# st & \# images \\ \hline
vortex~\cite{silver1997tracking}  & $128 \times 128 \times 128 \times 90$  & $812$ & $30$ & $24360$ \\

Tangaroa~\cite{popinet2004experimental}   & $300 \times 180 \times 120 \times 150$ & $812$ & $30$ & $24360$ \\

tornado~\cite{crawfis1993texture} & $128 \times 128 \times 128 \times 48$& $812$ &$30$ & $24360$ \\ 
\end{tabular}
}
\label{tab:dataset}
\end{table}
\vspace{-.1in} 
%===================================

\begin{table}[htb]
\caption{Average PSNR (dB) and LPIPS, BPP, and total ET (hours) DT (seconds). Each case has $24360$ images with a resolution of $1024 \times 1024$. The best ones are highlighted in bold.}
\vspace{-0.05in}
\centering
%x1.156
%{\scriptsize
%\resizebox{3.5in}{!}{
\resizebox{\linewidth}{!}{
\begin{tabular}{ccllllllllll}
& & \multicolumn{5}{|c}{IR images}  & \multicolumn{5}{|c}{DVR images}  \\ 
dataset             & method & \multicolumn{1}{|c}{PSNR$\uparrow$} & {LPIPS$\downarrow$} & {BPP$\downarrow$} & ET$\downarrow$ & DT$\downarrow$
                                          & \multicolumn{1}{|c}{PSNR$\uparrow$} & {LPIPS$\downarrow$} & {BPP$\downarrow$} & ET$\downarrow$ & DT$\downarrow$   \\ \hline
%\multirow{3}{*}{vortex}}    
                    &E-NeRV  & \multicolumn{1}{|l}{27.17} &0.1432  &0.0031 & 149.46 &1925.47
                    & \multicolumn{1}{|l}{21.77} & 0.1154 & 0.0031  &151.58 & 1939.19
                               \\
           &HNeRV   & \multicolumn{1}{|l}{21.15} & 0.2701 &\textbf{0.0019}  &72.72  & 391.25   
           & \multicolumn{1}{|l}{20.17} &0.2022 & \textbf{0.0019} &69.63 & 916.33 
                             \\
    vortex  &NeRVI   & \multicolumn{1}{|l}{26.80} &0.1386 &0.0356  &251.46 & 3965.49  
    & \multicolumn{1}{|l}{24.30} &0.0602 &0.0356  &255.56 & 2982.23 
                             \\
            &ECSIC   & \multicolumn{1}{|l}{36.27} &\textbf{0.0980} & 0.0915 &1.20 & \textbf{259.90}  
            & \multicolumn{1}{|l}{34.77} &0.0139 & 0.1437 &1.20 & \textbf{250.66} 
                             \\
                    &FCNR   & \multicolumn{1}{|l}{\textbf{37.47}} &0.1025 &0.0693  &\textbf{1.18} & 269.67  
                    & \multicolumn{1}{|l}{\textbf{34.85}} & \textbf{0.0132} & 0.1212 & \textbf{1.19} & 296.70  
                             \\
                     \hline
      %\multirow{3}{*}{vortex}}    
                    &E-NeRV  & \multicolumn{1}{|l}{25.96} &0.0093 & 0.0031 &147.52 & 4362.98  
                    & \multicolumn{1}{|l}{25.43} &0.1103 & 0.0031 &147.13 & 4203.47
                             \\
           &HNeRV  & \multicolumn{1}{|l}{23.91} &0.1690 &\textbf{0.0015} &57.07  &4872.01 
           & \multicolumn{1}{|l}{24.17} &0.1759  & \textbf{0.0015} & 71.50    & 1239.88 
            \\
   Tangaroa   &NeRVI   & \multicolumn{1}{|l}{28.16} &0.0750 &0.0356 & 181.13 & 2015.16  
   & \multicolumn{1}{|l}{26.39} &0.0964  &0.0359 & 247.33 & 2137.02
                             \\
                    &ECSIC   & \multicolumn{1}{|l}{37.82} &0.0149 &0.0895 & \textbf{1.18} & \textbf{306.94} 
                    & \multicolumn{1}{|l}{\textbf{34.61}} &\textbf{0.0153}  &0.1405 & 1.20 & 246.52
                                 \\
                        &FCNR   & \multicolumn{1}{|l}{\textbf{38.12}} &\textbf{0.0145}  &0.0709 & \textbf{1.18} & 319.84 
                        & \multicolumn{1}{|l}{34.45} &0.0177  &0.1109 & \textbf{1.17} & \textbf{211.73}
                             \\ \hline
                    &E-NeRV  & \multicolumn{1}{|l}{36.72} &0.1389  &0.0031 & 50.14 & 4643.82 
                    & \multicolumn{1}{|l}{35.09} &0.0038  &0.0031 & 50.34  & 5157.63 
                             \\
           &HNeRV   & \multicolumn{1}{|l}{34.53} &0.0544 & \textbf{0.0015} &28.54  &5359.20 
           & \multicolumn{1}{|l}{31.97} &0.0506  &\textbf{0.0016} & 30.75  & 1804.20
                              \\
    tornado  &NeRVI   &\multicolumn{1}{|l}{\textbf{38.21}} &\textbf{0.0359}  &0.0356 & 90.56  &  9609.40 
    & \multicolumn{1}{|l}{36.27} &0.0336  &0.0356 & 88.99  & 2030.20 
                              \\
                    &ECSIC   & \multicolumn{1}{|l}{36.30} &0.0700  &0.0580 & 0.40 & \textbf{231.71} 
                    & \multicolumn{1}{|l}{36.51} &0.0501  &0.0982 & 0.42  & \textbf{180.17}
                             \\
                    &FCNR   & \multicolumn{1}{|l}{38.07} &0.0685 &0.0280 & \textbf{0.39} & 326.91 
                    & \multicolumn{1}{|l}{\textbf{37.35}} &\textbf{0.0386}  &0.0359 & \textbf{0.39} & 352.00
                             \\
\end{tabular}
}
%}
\label{tab:comp-result-iso-volume-more}
\end{table}
%\vspace{-.1in} 

{\bf Loss functions.}
Image compression models can be optimized for a weighted sum of the rate and distortion losses~\cite{balle2017end}
\begin{equation}
L_{\rm total}=L_R+\lambda L_D,
\end{equation}
where $L_R$ is the rate loss, $L_D$ is the distortion loss, and $\lambda \in \mathbb{R}$ is a trade-off weight. 
$L_D$ is the expectation of the mean squared errors between the input and reconstructed images
\begin{equation}
L_D=\mathbb{E}_{x_l, x_r \sim p_x}\big[\left\|x_l-\hat{x}_l\right\|^2_2+\left\|x_r-\hat{x}_r\right\|^2_2\big].
\end{equation}
Following the rate loss in ECSIC, $L_R$ is the expected sum of the cross entropy between the predicted distribution of our entropy model and the true distribution of the latents or hyper-latents
\begin{align}
    \begin{split}
        L_R = \mathbb{E}_{x_l, x_r \sim p_{x}} \big[
        &-\log_2 p( \hat{z}_l\mid\phi^z_l)\\
        &-\log_2 p( \hat{z}_r\mid\Phi_{{\rm cont}_z},\phi^z_r, \hat{z}_l)\\
        &-\log_2 p( \hat{y}_l\mid\Phi_{h_D}, \hat{z}_r, \hat{z}_l)\\
        &-\log_2 p( \hat{y}_r\mid\Phi_{{\rm cont}_y}, \Phi_{h_D}, \hat{y}_l, \hat{z}_r, \hat{z}_l) \big],
    \end{split}
\end{align}
where $\Phi_{{\rm cont}_z}, \Phi_{{\rm cont}_y}, \Phi_{h_D}$ denote the parameters of ${\rm cont}_z, {\rm cont}_y,$ and $h_D$, respectively.

\begin{figure}[htb]
%\vspace{-.1in} 
  \begin{center}
  $\begin{array}{c@{\hspace{0.01in}}c@{\hspace{0.01in}}c@{\hspace{0.01in}}c@{\hspace{0.01in}}c@{\hspace{0.01in}}c@{\hspace{0.01in}}c}
    \includegraphics[width=0.16\linewidth]{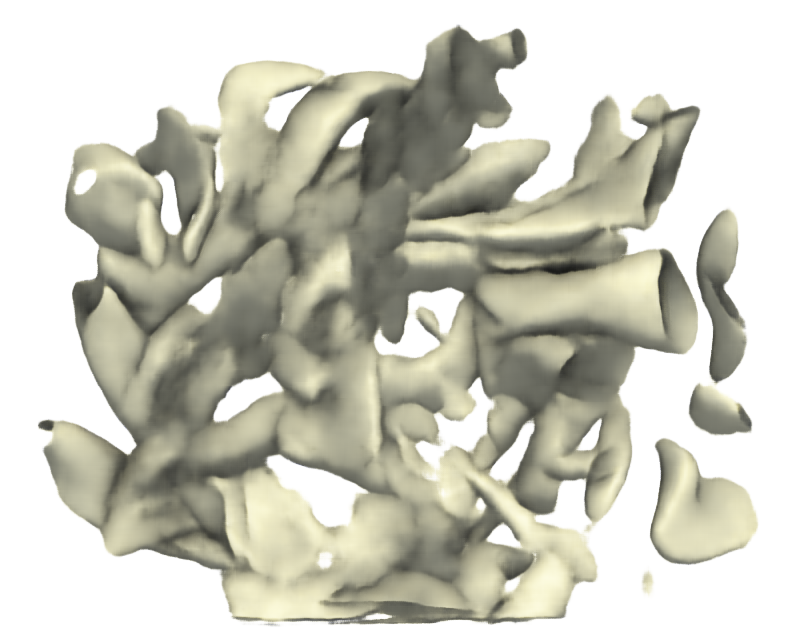}&
    \includegraphics[width=0.16\linewidth]{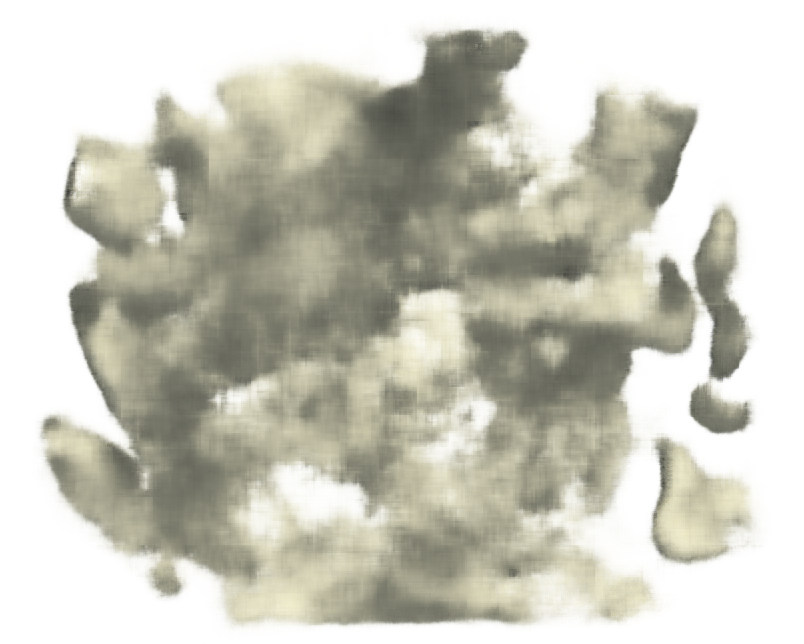}&
    \includegraphics[width=0.16\linewidth]{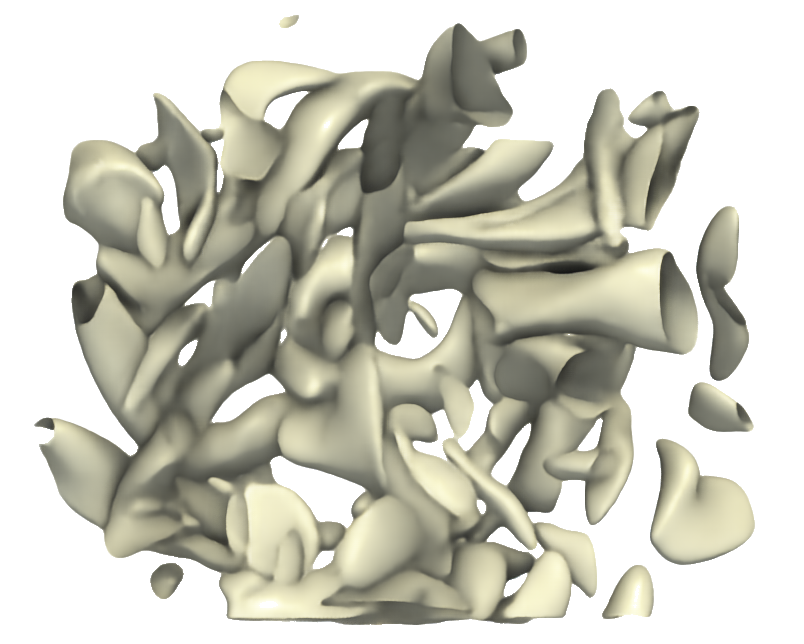}&
    \includegraphics[width=0.16\linewidth]{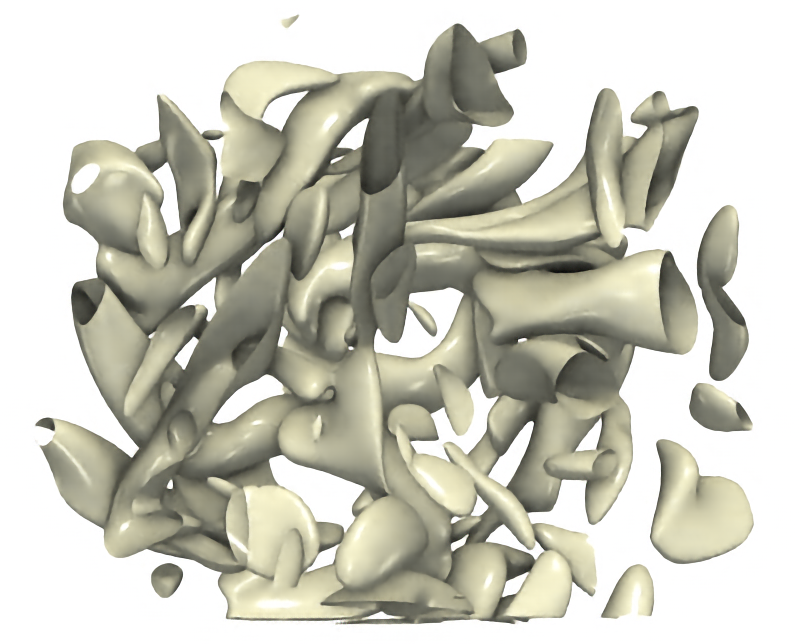}&
    \includegraphics[width=0.16\linewidth]{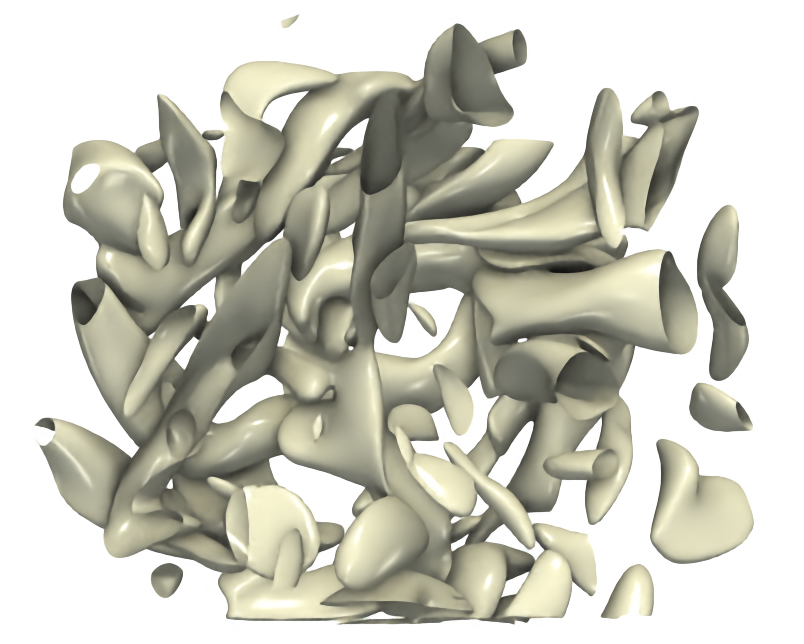}&
    \includegraphics[width=0.16\linewidth]{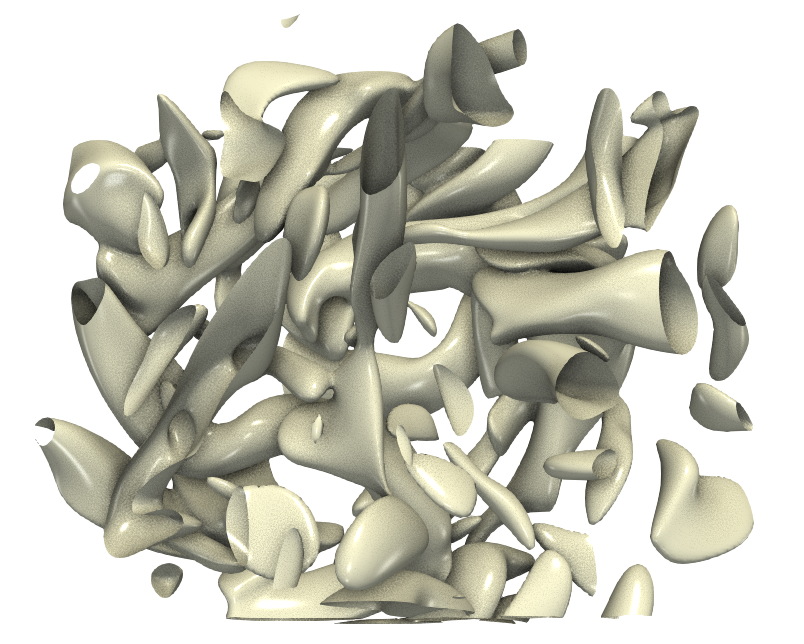}\\
    \includegraphics[width=0.16\linewidth]{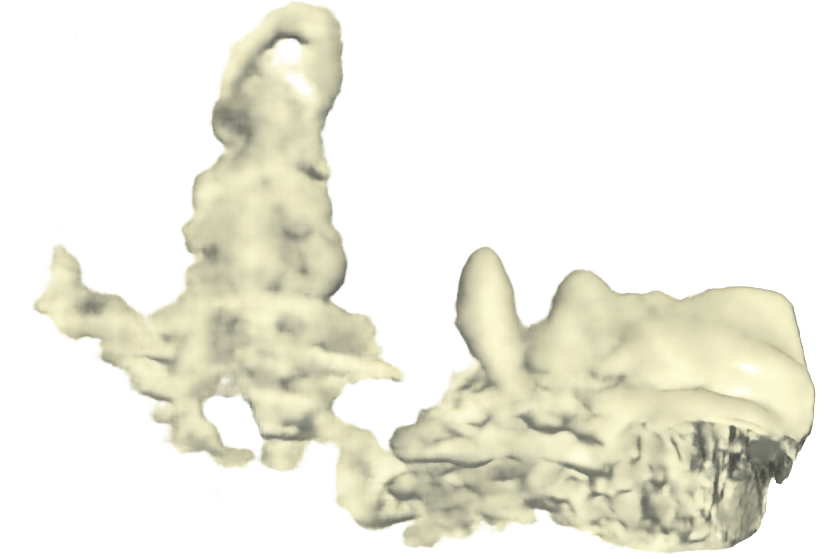}&
    \includegraphics[width=0.16\linewidth]{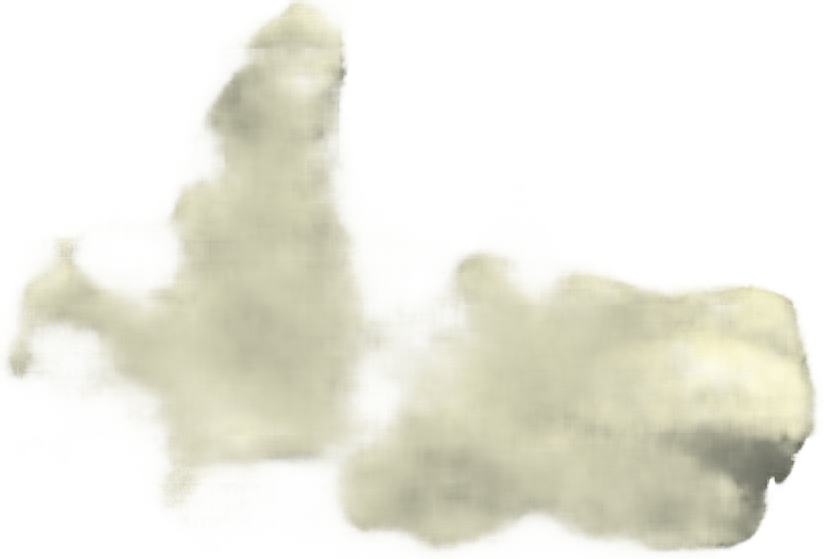}&
    \includegraphics[width=0.16\linewidth]{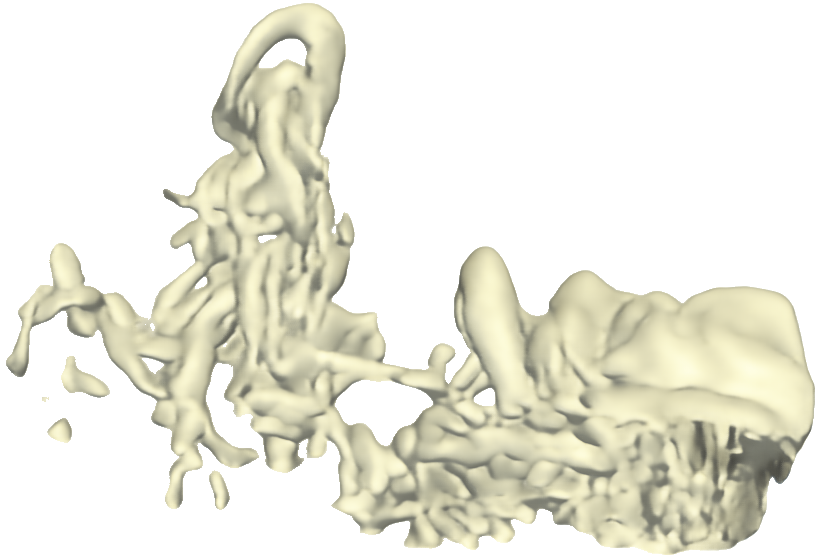}&
    \includegraphics[width=0.16\linewidth]{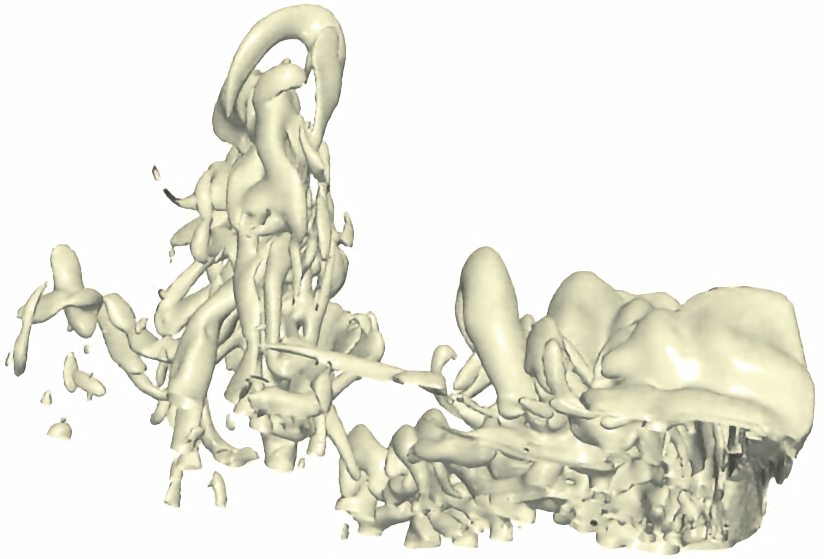}&
    \includegraphics[width=0.16\linewidth]{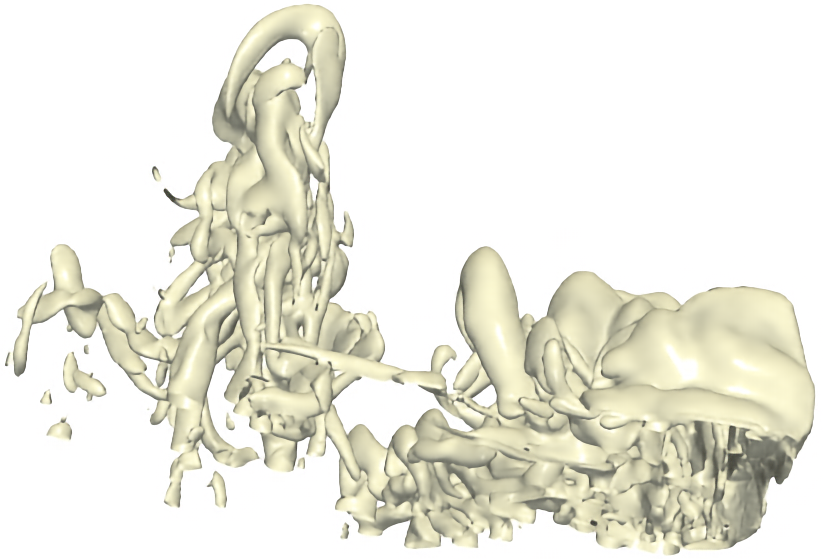}&
    \includegraphics[width=0.16\linewidth]{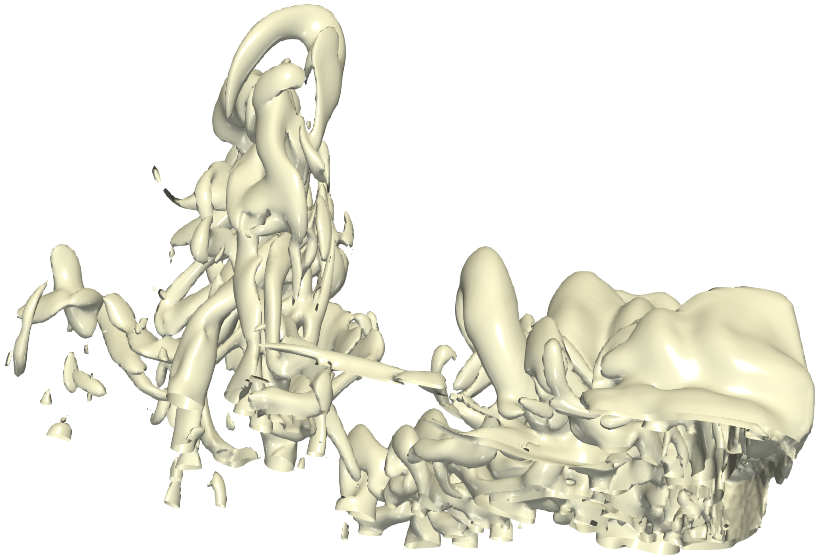}\\
    \includegraphics[width=0.16\linewidth]{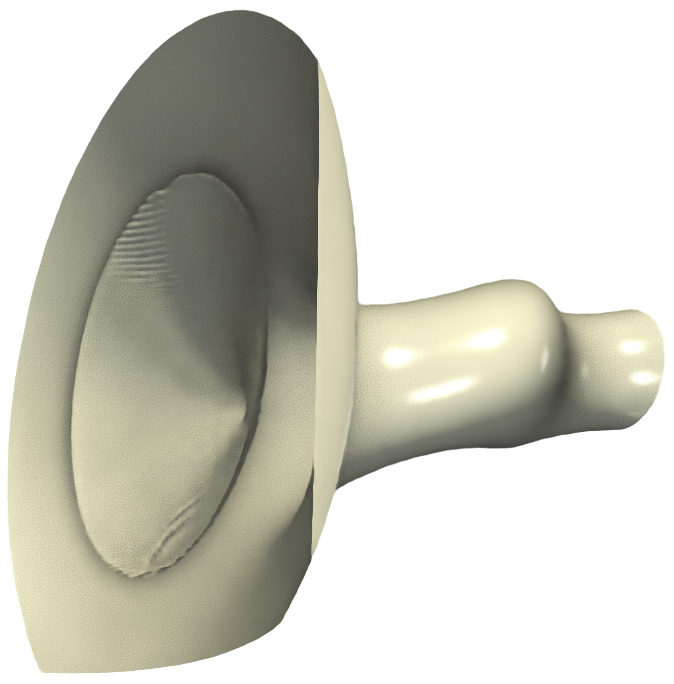}&
    \includegraphics[width=0.16\linewidth]{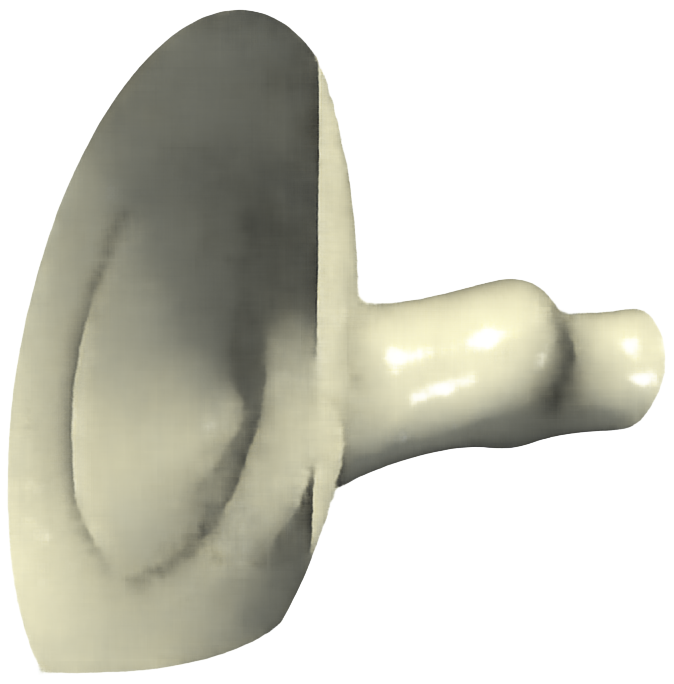}&
    \includegraphics[width=0.16\linewidth]{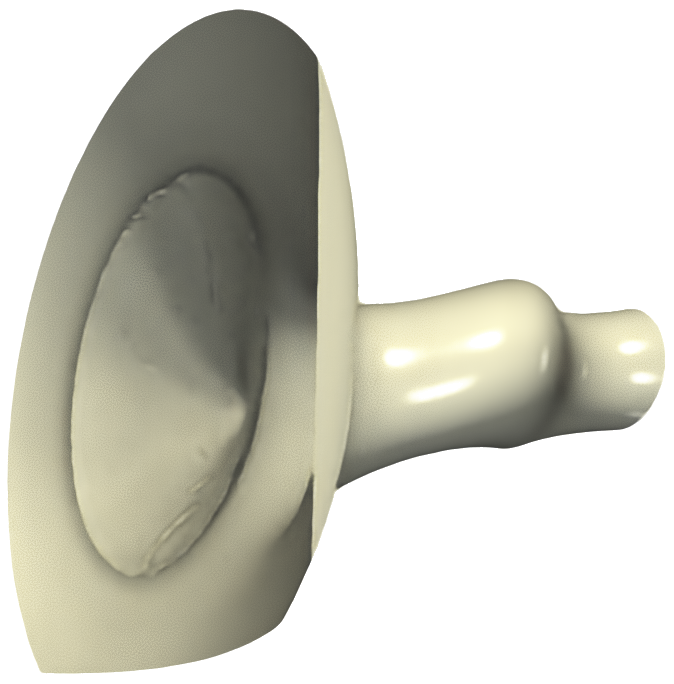}&
    \includegraphics[width=0.16\linewidth]{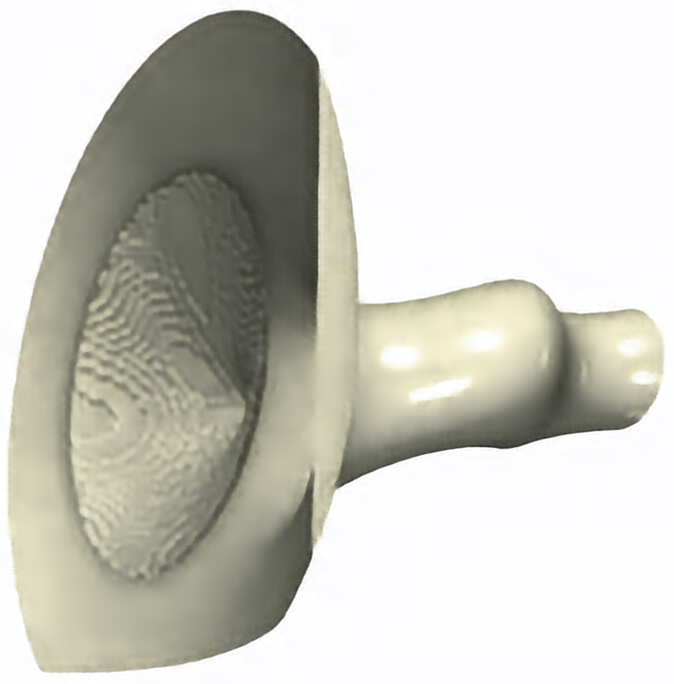}&
    \includegraphics[width=0.16\linewidth]{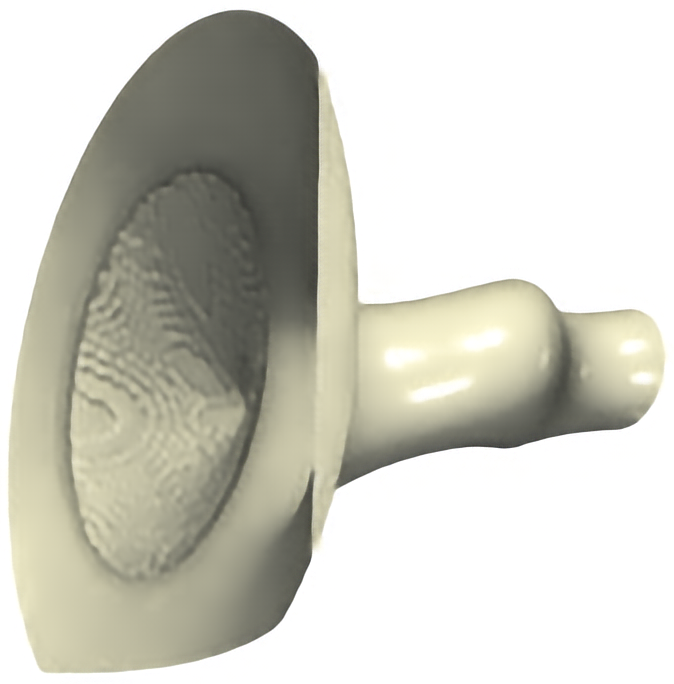}&
    \includegraphics[width=0.16\linewidth]{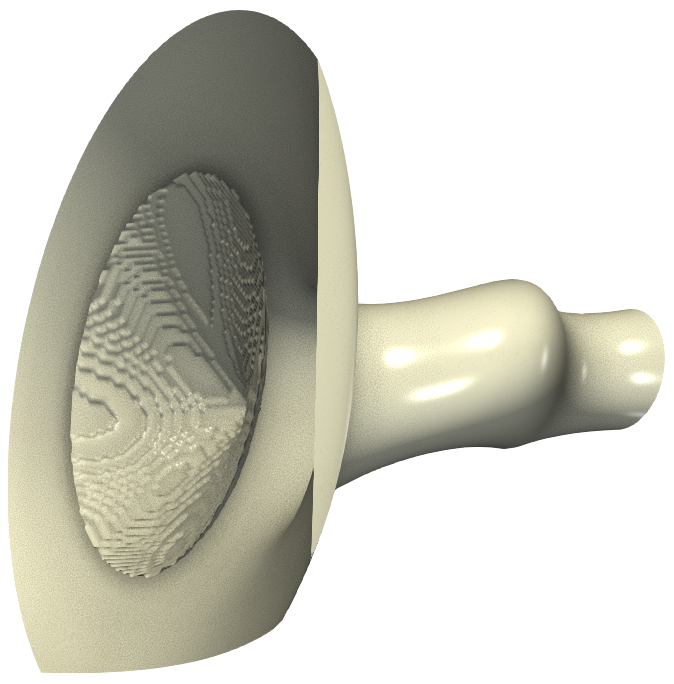}\\
    \includegraphics[width=0.16\linewidth]{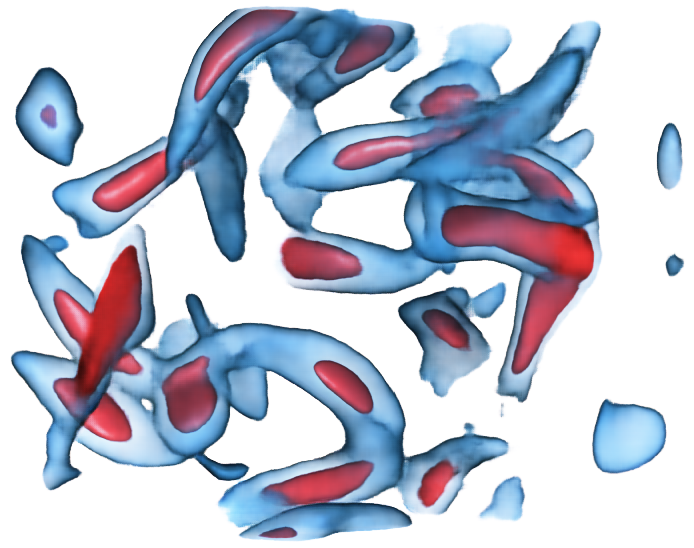}&
    \includegraphics[width=0.16\linewidth]{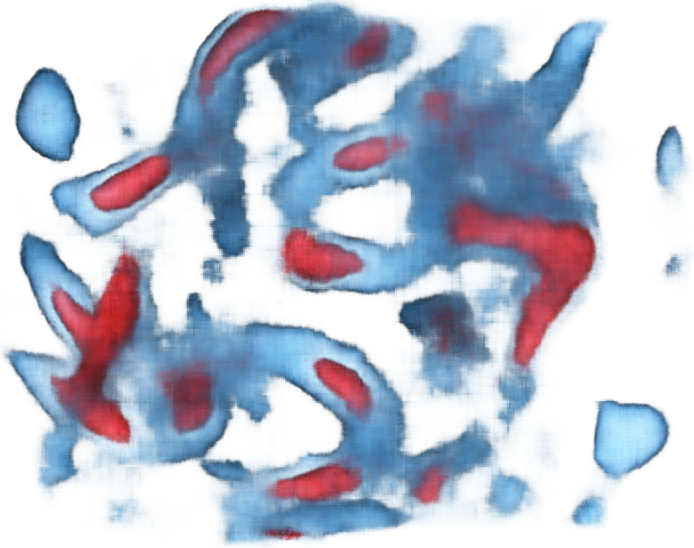}&
    \includegraphics[width=0.16\linewidth]{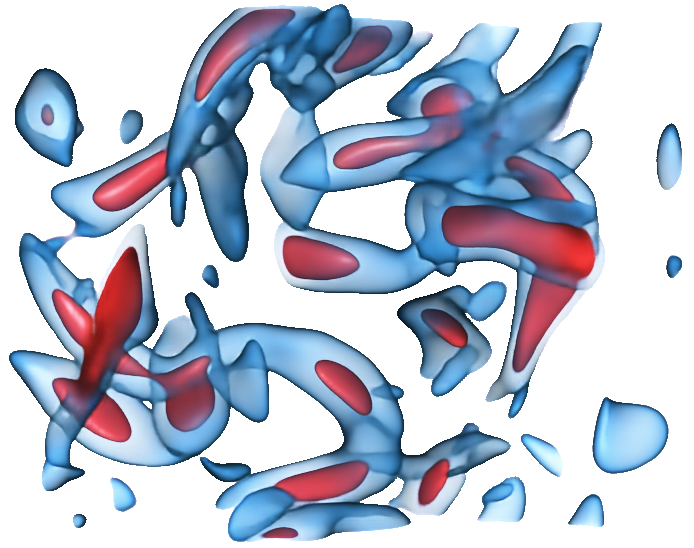}&
    \includegraphics[width=0.16\linewidth]{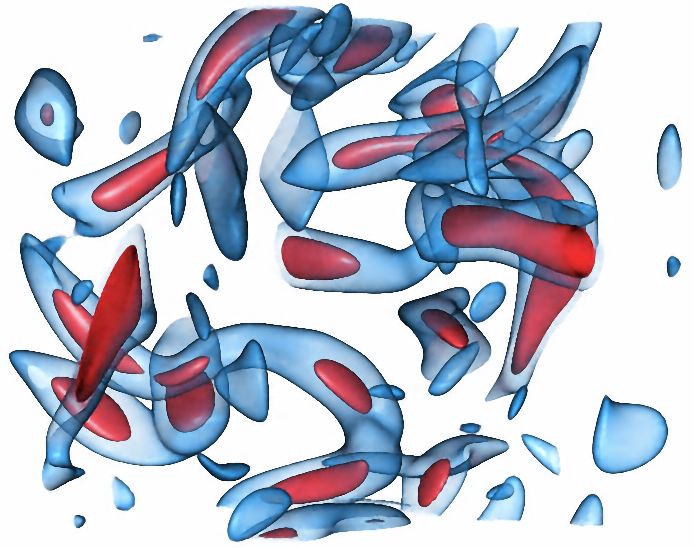}&
    \includegraphics[width=0.16\linewidth]{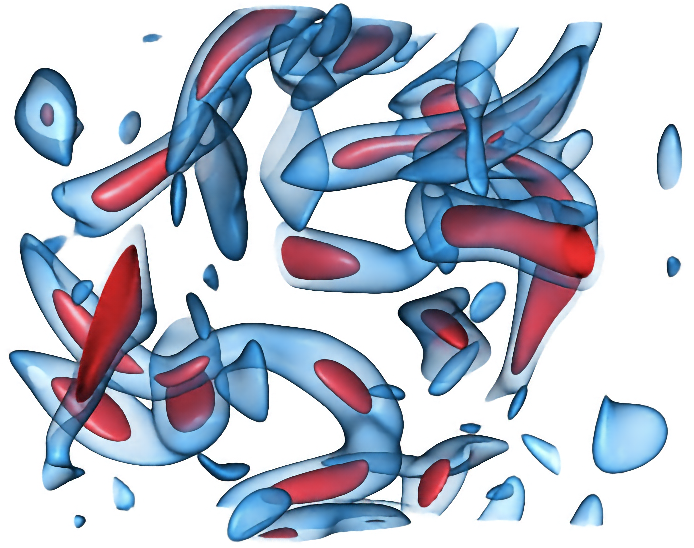}&
    \includegraphics[width=0.16\linewidth]{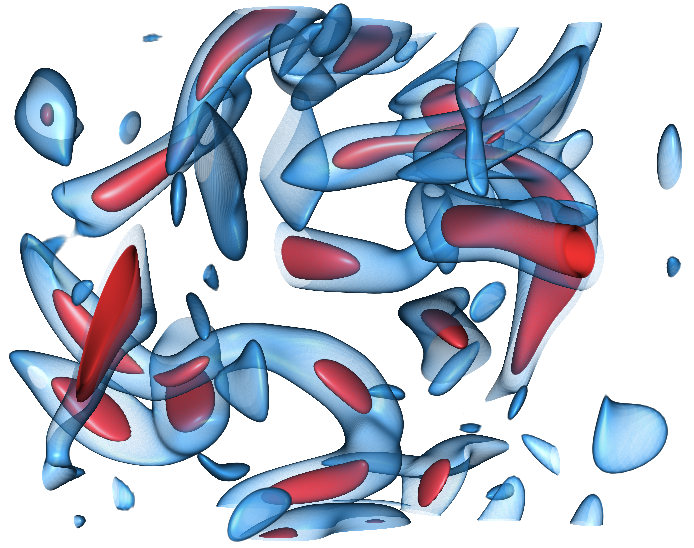}\\
    \includegraphics[width=0.16\linewidth]{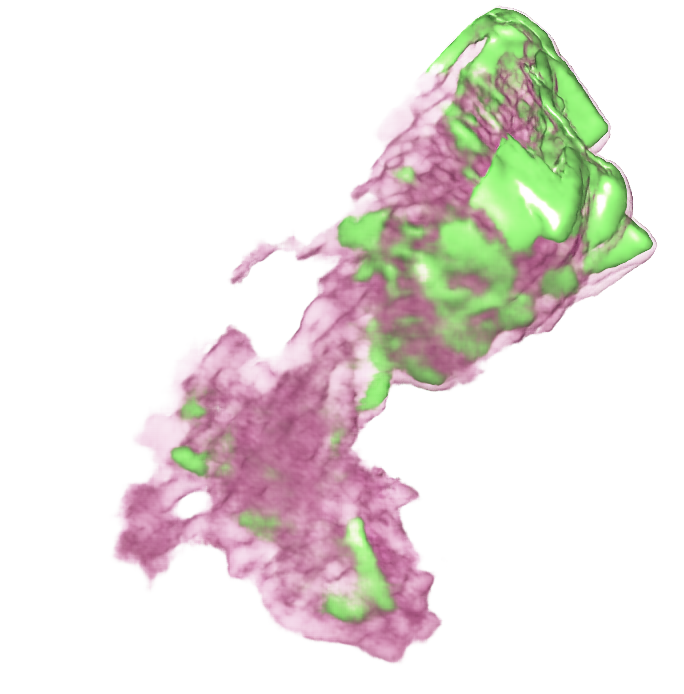}&
    \includegraphics[width=0.16\linewidth]{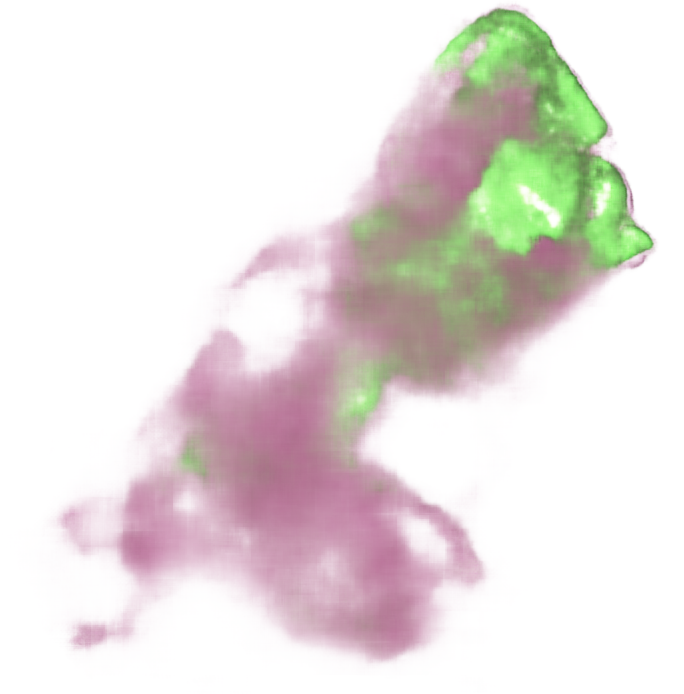}&
    \includegraphics[width=0.16\linewidth]{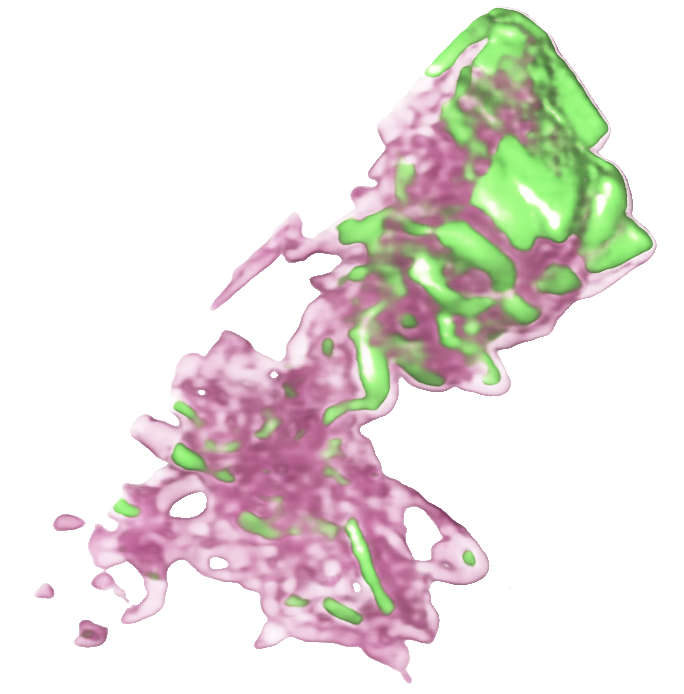}&
    \includegraphics[width=0.16\linewidth]{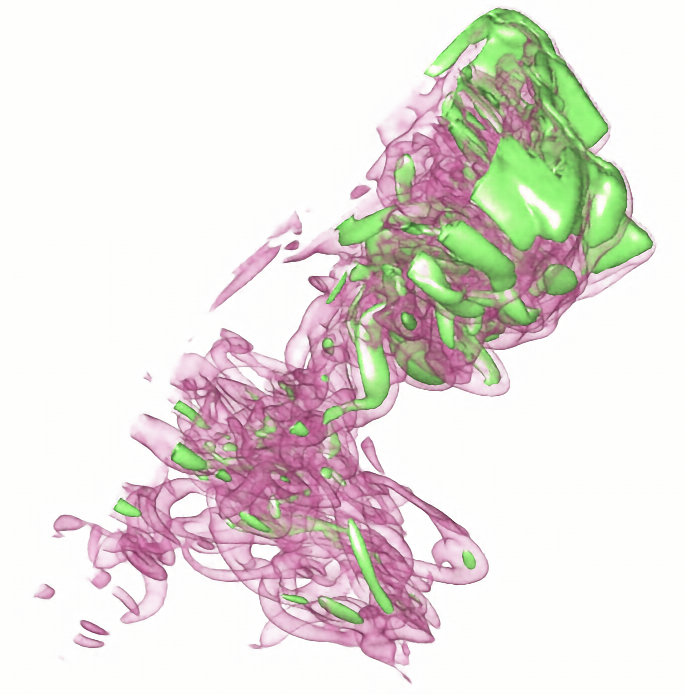}&
    \includegraphics[width=0.16\linewidth]{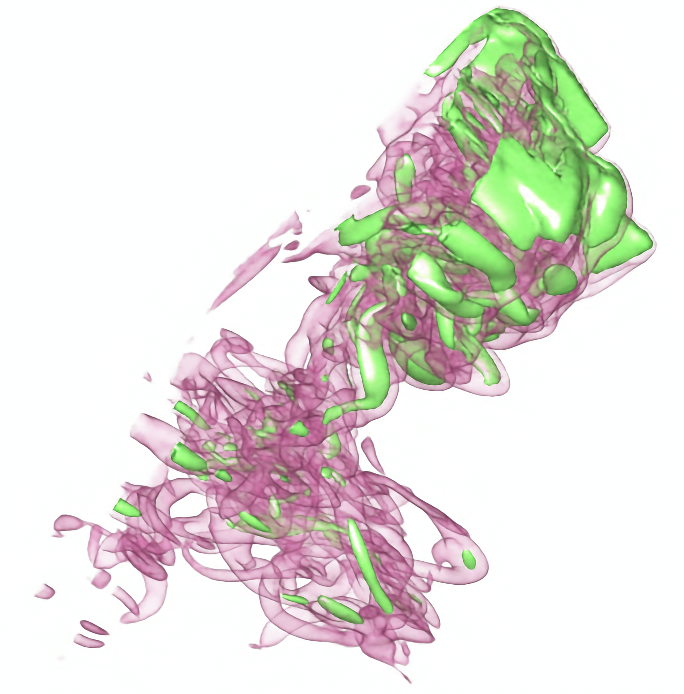}&
    \includegraphics[width=0.16\linewidth]{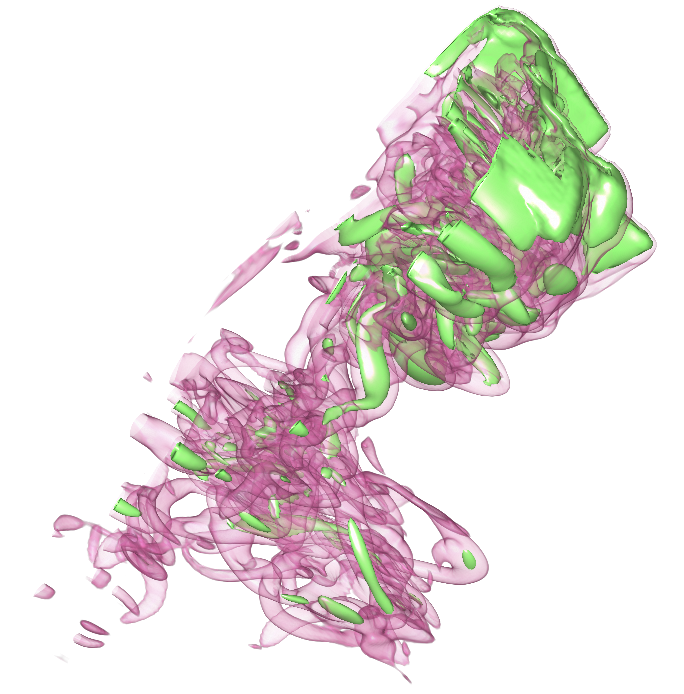}\\
    \includegraphics[width=0.16\linewidth]{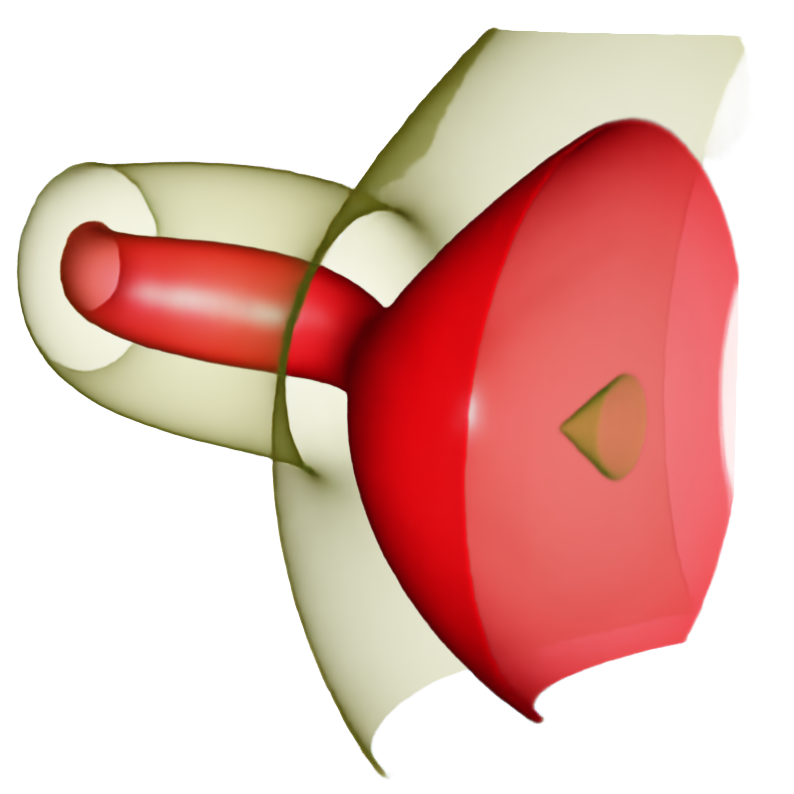}&
    \includegraphics[width=0.16\linewidth]{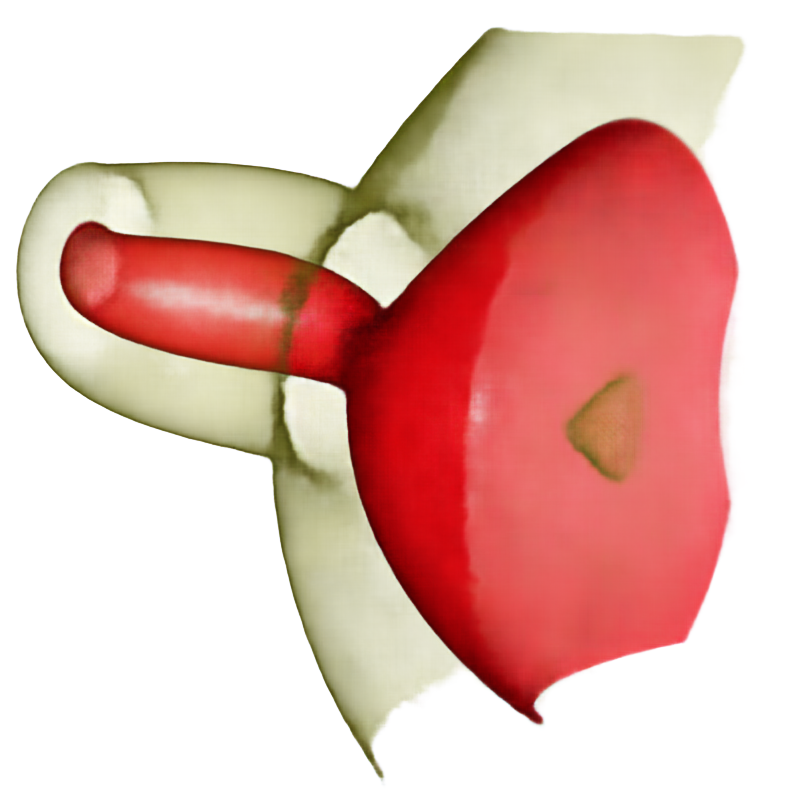}&
    \includegraphics[width=0.16\linewidth]{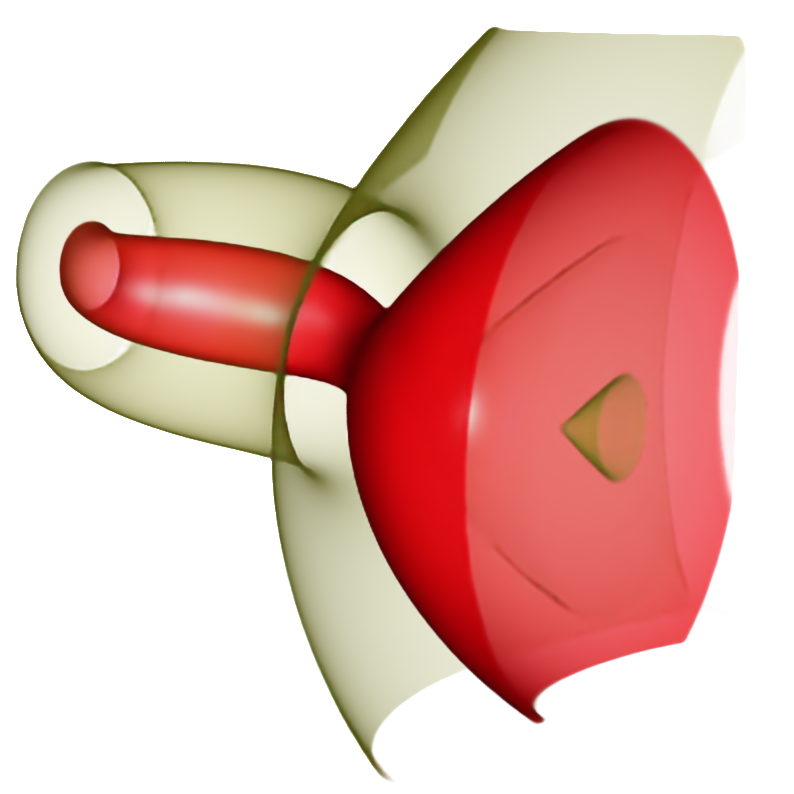}&
    \includegraphics[width=0.16\linewidth]{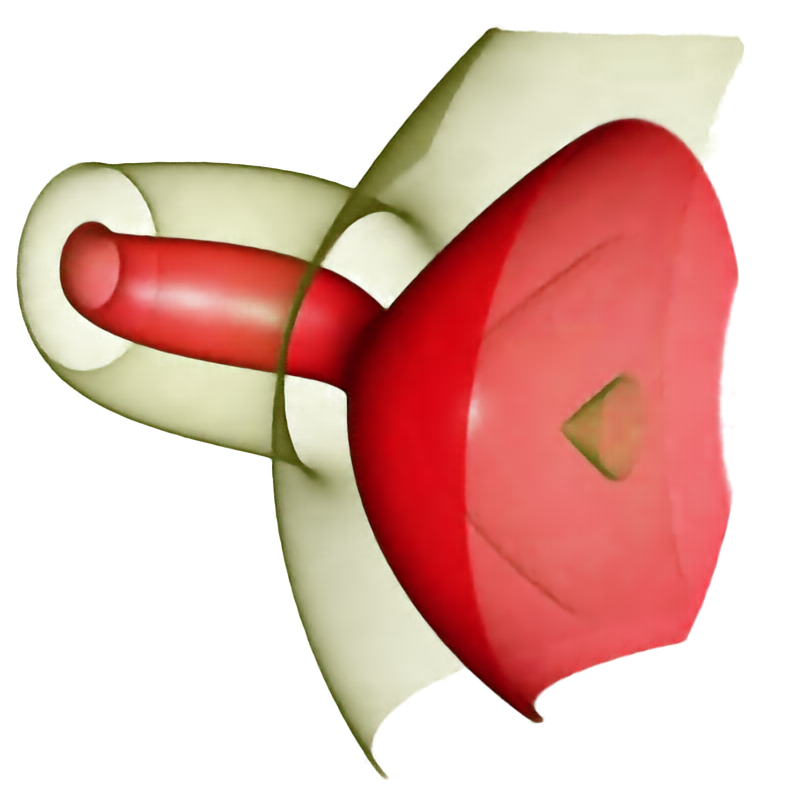}&
    \includegraphics[width=0.16\linewidth]{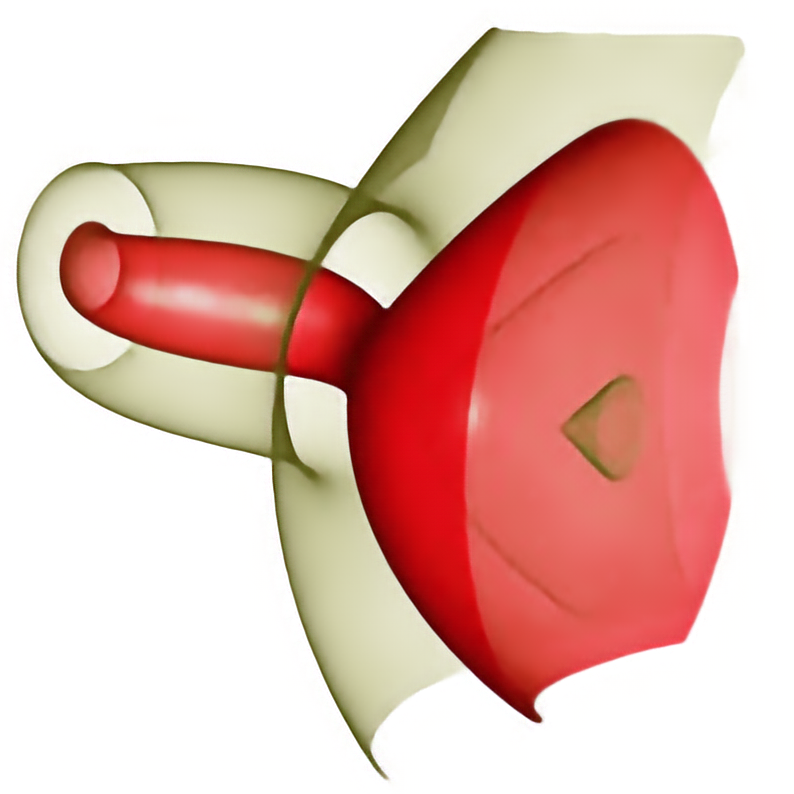}&
    \includegraphics[width=0.16\linewidth]{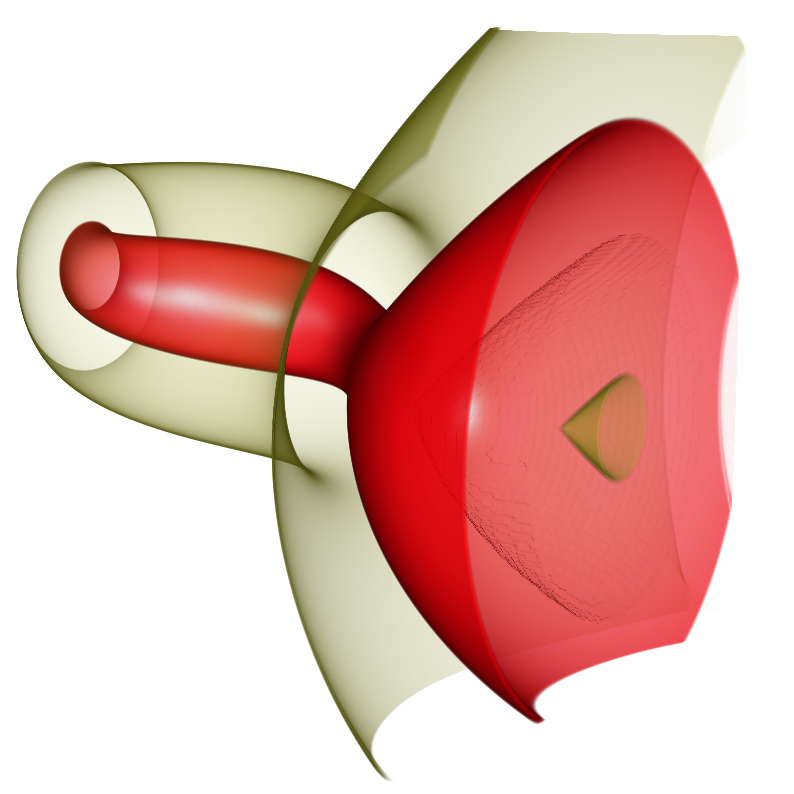}\\
    \mbox{\footnotesize (a) E-NeRV} & \mbox{\footnotesize (b) HNeRV} & \mbox{\footnotesize (c) NeRVI} & \mbox{\footnotesize (d) ECSIC} & \mbox{\footnotesize (e) FCNR} & \mbox{\footnotesize (f) GT} 
  \end{array}$
 \end{center}
\vspace{-.25in} 
 \caption{Decompressed IR and DVR images. The datasets are vortex, Tangaroa, and tornado, respectively.} 
 \label{fig:baselines-ir-dvr}
\end{figure}
%\vspace{-.1in} 

\vspace{-0.075in}
\section{Results and Discussion}

{\bf Datasets, setting, and training.}
Table~\ref{tab:dataset} shows the three datasets we experimented with. 
We picked 30 consecutive timesteps for each dataset. 
To ensure an even distribution of viewpoints across the volume data, we determined camera positions using the vertices of an icosphere, approximating a sphere using equilateral triangles. 
We selected a subdivision level resulting in 812 vertices to generate the training set for each timestep, producing a good quantity of images for compression. 
%For each viewpoint, we log the camera position in the world coordinate system for 3D-aware view synthesis methods, and then convert it to azimuth (ranging from -180 to 180) and elevation (ranging from -90 to 90) for 2D-based approaches. 
The image resolutions were all set to 1024$\times$1024. 
For ECSIC and FCNR, which require image pairs, we sorted the images at each timestep first by $\theta$ and then by $\varphi$ if there is a tie. 
We then chose the image with an even index $j$ (starting from 0) as the left image and the image with index $j+1$ as its right counterpart. 
Since ECSIC and FCNR possess generalization ability, we decreased the number of training images to $1/6$ (evenly selected $1/2$ of the sampled views and $1/3$ of the timesteps), i.e., 4060 images. We evaluated these two models on all 24360 images during inference. 
%We reduced the number of sampled views at each timestep by half, only selecting the image pairs with an even index as training inputs. We further chose timesteps that are multiples of 3 to cut down the timesteps by $1/3$.

%{\bf Implementation Details.}
We implemented FCNR using PyTorch. We chose the number of channels in the encoding and decoding modules to be 192, the number of latent channels to be 48, and the number of attention heads in JCTM to be 2. All experiments were run on an NVIDIA A40 GPU. 
Adam optimizer was utilized for gradient descent ($\beta_1$=0.9, $\beta_2$=0.999), and the learning rate was set as $10^{-4}$. We trained FCNR with a batch size of 1. The number of training epochs was 3 for the vortex and Tangaroa datasets and 1 for the tornado dataset.

{\bf Baselines and evaluation metrics.}
We compared FCNR with three state-of-the-art INR-based methods, including E-NeRV~\cite{li2022enerv}, HNeRV~\cite{chen2023hnerv}, and NeRVI~\cite{Gu-CG23}, and one stereo image compression method, ECSIC~\cite{wodlinger2024ecsic}. We extended E-NeRV by feeding all $(t, \theta, \varphi)$ to the model, first normalized to $[0,1]$ and then 
%positional encoded as the network input" 
input to the network after PE. All INR-based methods were trained until convergence for a fair comparison. The number of training epochs for ECSIC was the same as FCNR for each dataset. 
For quantitative evaluation in the image space, we employed peak signal-to-noise ratio (PSNR) and learned perceptual image patch similarity (LPIPS)~\cite{Zhang-CVPR18}. Besides, encoding time (ET) and decoding time (DT) are also recorded.

\begin{table}[htb]
\caption{Average PSNR (dB) and LPIPS, BPP, and total ET (hours) DT (seconds) of ECSIC, its varying modifications, and FCNR on the tornado IR dataset.}
\vspace{-0.05in}
\centering
%{\scriptsize
\resizebox{2.5in}{!}{
%\resizebox{\columnwidth}{!}{
\begin{tabular}{c|ccccc}
             method  & PSNR$\uparrow$ & LPIPS$\downarrow$ &BPP$\downarrow$ &ET$\downarrow$ &DT$\downarrow$   \\ \hline
%\multirow{3}{*}{vortex}} 
 			ECSIC  & 36.30    & 0.0700    & 0.0580  & 0.40 & 231.71 \\   
				JCT-Only  & 37.81    & \textbf{0.0685}    & 0.0666  & \textbf{0.39} & 270.15 \\   
		    PE-Only  & 37.19& 0.0708  & 0.0370  & 0.40 & \textbf{228.91} \\
                FCNR  &\textbf{38.07}   & \textbf{0.0685}   & \textbf{0.0280}  & \textbf{0.39} & 326.91 \\
\end{tabular}
%}
}
\label{tab:ablation-study}
\end{table}
%\vspace{-.1in} 

\begin{comment}
\begin{table}[htb]
\caption{Comparison of different methods over encoding speed (ES), decoding speed (DS), compression ratio (CR), and image quality (IQ).}
%\vspace{-0.1in}
\centering
%{\scriptsize
\resizebox{2.5in}{!}{
%\resizebox{\columnwidth}{!}{
\begin{tabular}{c|cccc}
             method  & ES & DS &CR & IQ   \\ \hline
%\multirow{3}{*}{vortex}} 
 			E-NeRV  & very slow    & very slow    &  very high  & low \\   
				HNeRV  & slow    & slow    &  very high  & very low \\   
		      NeRVI  & very slow    & very slow    &  high  & medium \\
                ECSIC  & very fast   & fast   & medium  & high\\
                FCNR   & very fast   & fast   & medium  & high \\
\end{tabular}
%}
}
\label{tab:comp-methods}
\end{table}
\end{comment}

\begin{figure}[htb]
%\vspace{-.1in} 
  \begin{center}
    \includegraphics[width=1.\linewidth]{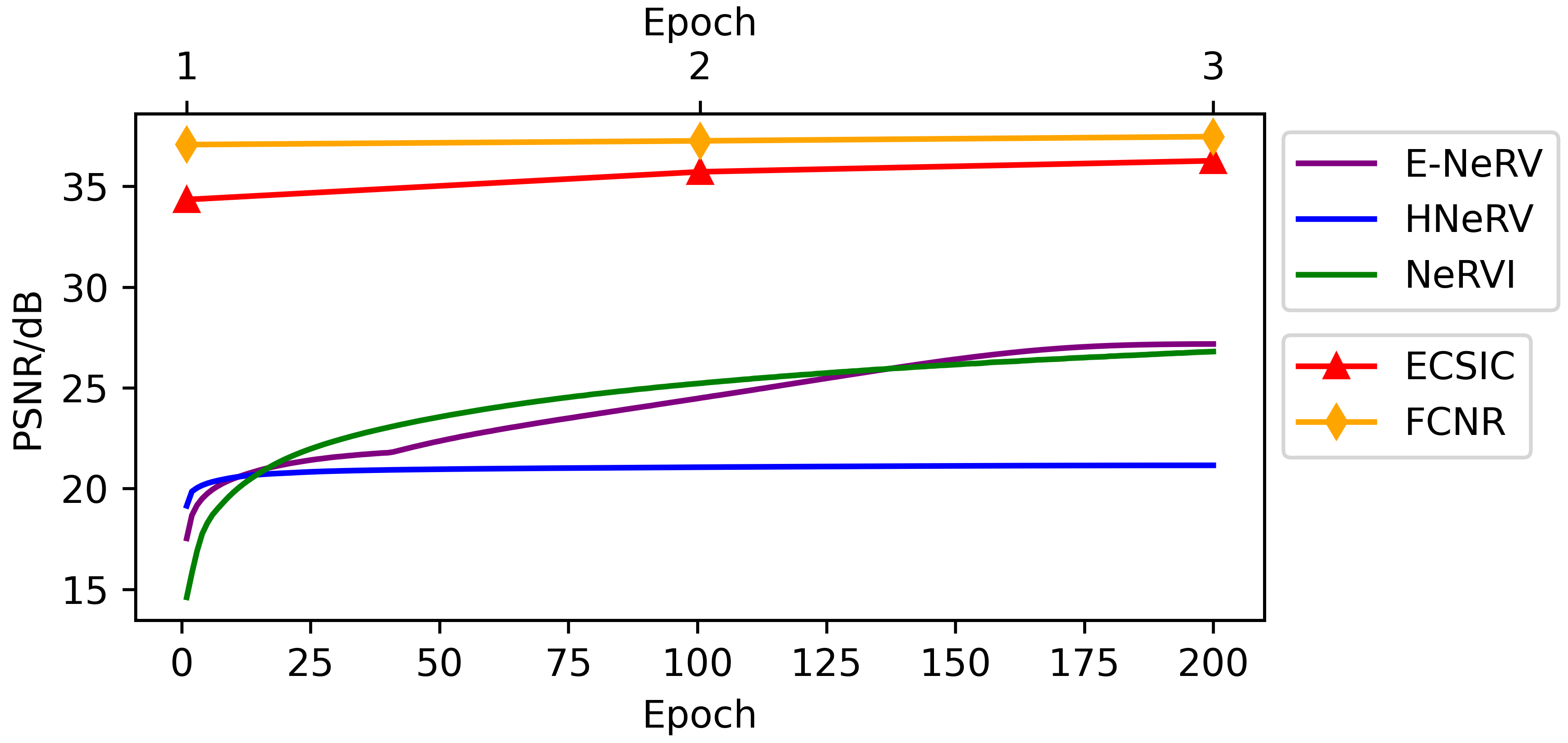}
 \end{center}
\vspace{-.25in}
 \caption{PSNR comparison of all methods on the vortex IR dataset. E-NeRV, HNeRV, and NeRVI were all trained for 200 epochs. Both ECSIC and FCNR were trained for 3 epochs.} %associated with visualization parameters $(t_l,\theta_l,\varphi_l)$ and $(t_r,\theta_r,\varphi_r)$
 \label{fig:psnr-curve}
\end{figure}

{\bf Results.}
Table~\ref{tab:comp-result-iso-volume-more} compares FCNR with state-of-the-art baselines quantitatively. Figure~\ref{fig:baselines-ir-dvr} shows the decompressed rendering images for all datasets under chosen $(t, \theta, \varphi)$. 
All images are cropped for closer comparison. 

FCNR achieves the highest PSNR and lowest LPIPS for most cases, even on images unseen during training, demonstrating its interpolation ability. Although INR-based methods (E-NeRV, HNeRV, and NeRVI) lead to lower BPP, they have limited reconstruction quality as depicted in Figure~\ref{fig:baselines-ir-dvr}. The vortex and Tangaroa datasets vary greatly at different timesteps, posing greater reconstruction challenges. HNeRV produces the most blurry images. The results of E-NeRV and NeRVI are much clearer. However, they still yield artifacts and distortions when multiple components overlap in rendering images and miss some components when images become complex.
For the tornado dataset, all methods generate clear images, but FCNR generates images with better PSNR and LPIPS than E-NeRV and HNeRV. Though NeRVI achieves the highest PSNR and the lowest LPIPS in the tornado IR dataset, it fails to reconstruct high-frequency details as shown in Figure~\ref{fig:baselines-ir-dvr}. By contrast, FCNR generates the best visual quality images, with the clearest high-frequency details and the fewest artifacts.

Moreover, the significant encoding time (48.36$\times$ to 232.21$\times$) and decoding time (1.45$\times$ to 29.39$\times$) over FCNR put INR-based baselines at a disadvantage. This is because E-NeRV and NeRVI need substantial training and more parameters to restore the lost information and reconstruct images from the rather limited and low-dimensional input and the necessity of learning both input embedding and decoder weights in HNeRV leads to a more complex design. In contrast, like ECSIC, FCNR drastically enhances encoding and decoding speed by fully utilizing the images for direct compression and reconstruction and exploiting mutual information between them.
%18.56-173.54
Compared with ECSIC, FCNR achieves very close, high-quality results for all datasets, with slight improvements in PSNR and LPIPS for the majority of cases. While ECSIC performs similarly to FCNR in image quality and encoding and decoding speed, its BPP is from 18.56\% (vortex DVR dataset) to 173.54\% (tornado DVR dataset) higher than FCNR across all cases. Such improvements make FCNR stand out from ECSIC.

Figure~\ref{fig:psnr-curve} compares FCNR and baseline methods in PSNR during training on the vortex IR dataset. It shows that all methods have been trained until convergence to ensure a fair comparison. The comparison highlights the effectiveness and efficiency of FCNR.

%These results demonstrate the effectiveness of JCTM and injecting visualization parameters to stereo context modules, enabling FCNR to capture more global correlation and diverse information, further improving the storage efficiency of encoded bitstreams.

%{\bf Qualitative results.}
%
%The vortex and Tangaroa datasets vary significantly at different timesteps, posting greater reconstruction difficulty. HNeRV produces the most blurry images. The results of E-NeRV and NeRVI are much clearer. However, they still yield blurry artifacts and distortions when multiple components overlap in rendering images and miss some parts when images become complex. ECSIC and FCNR achieve very close, high-quality results for all datasets, which are hard to differentiate.
%
%For the tornado dataset, all methods generate clear images, but FCNR generates images with the best visual quality and the fewest artifacts. Though NeRVI}

{\bf Ablation study.}
We performed an ablation study on the tornado IR dataset to show FCNR's differences from ECSIC. 
We compared FCNR and ECSIC with two architectural modifications: JCT-only and PE-only. 
For the JCT-only case, we modified ECSIC by changing all SCAMs to JCTMs. 
For the PE-only case, we extended ECSIC by transforming $(t_l, \theta_l, \varphi_l)$ to the input $\psi_l^z$ of $x_l$'s AD and $(t_r, \theta_r, \varphi_r)$ to the corresponding parameter $\phi_r^z$ of ${\rm cont}_z$ through PE and MLP.

\begin{figure}[htb]
%\vspace{-.1in} 
  \begin{center}
  $\begin{array}{c@{\hspace{0.0in}}c@{\hspace{0.0in}}c@{\hspace{0.0in}}c@{\hspace{0.0in}}c@{\hspace{0.0in}}c}
    \includegraphics[width=0.192\linewidth]{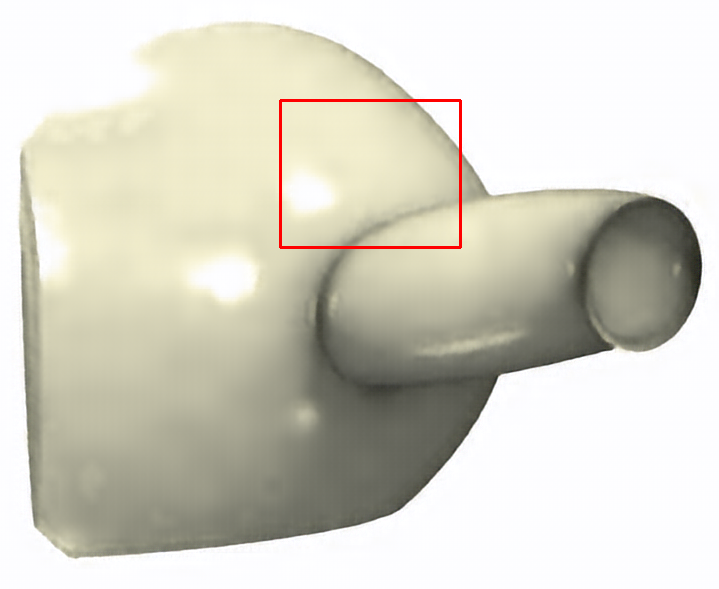}&
    \includegraphics[width=0.192\linewidth]{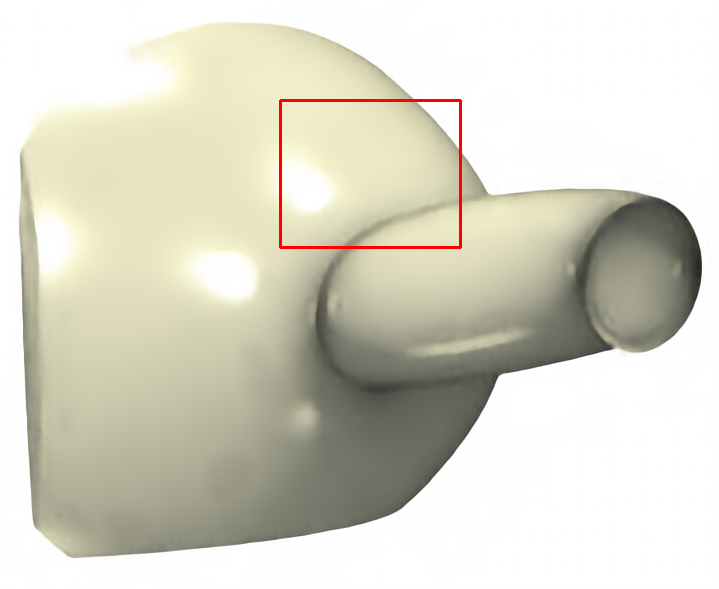}&
    \includegraphics[width=0.192\linewidth]{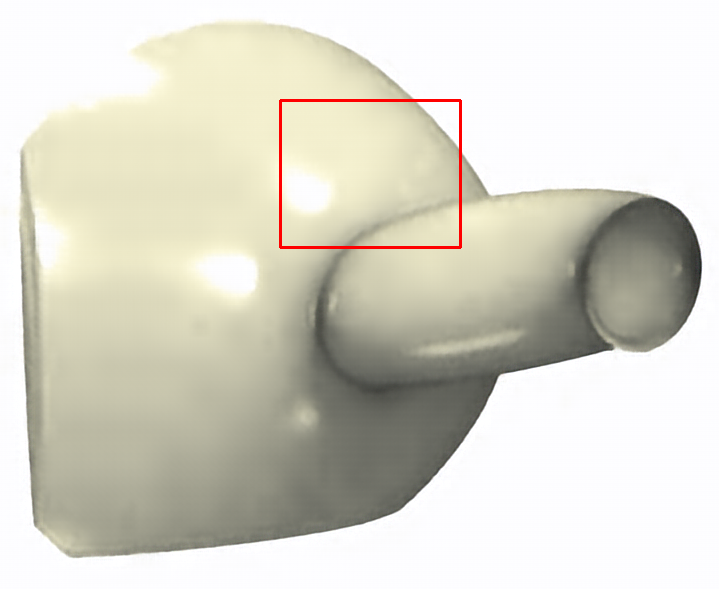}&
    \includegraphics[width=0.192\linewidth]{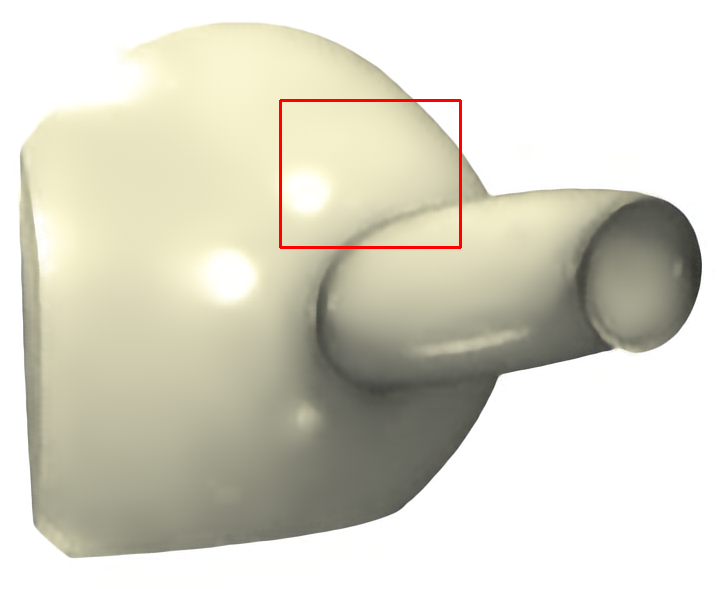}&
    \includegraphics[width=0.192\linewidth]{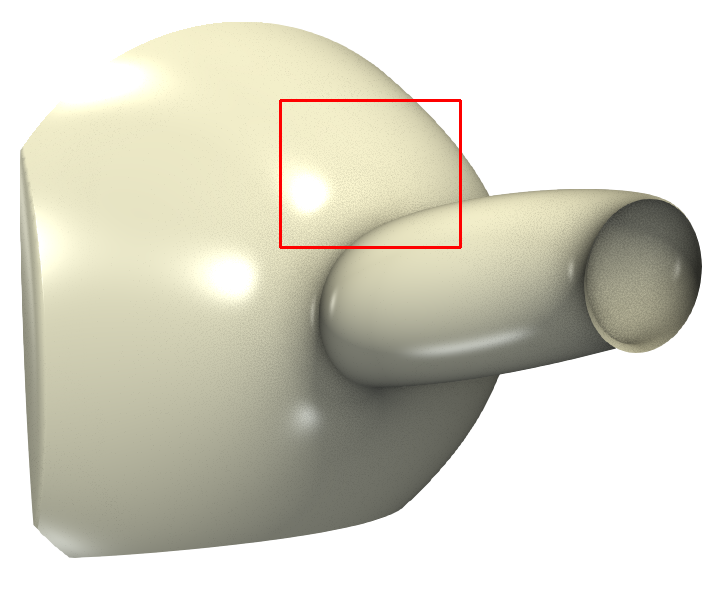}&\\
    \includegraphics[width=0.192\linewidth]{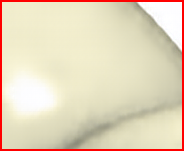}&
    \includegraphics[width=0.192\linewidth]{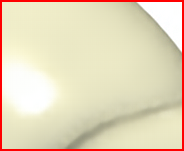}&
    \includegraphics[width=0.192\linewidth]{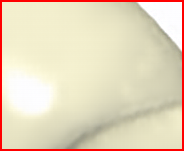}&
    \includegraphics[width=0.192\linewidth]{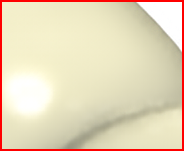}&
    \includegraphics[width=0.192\linewidth]{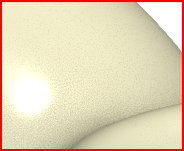}&
    \\
    \mbox{\footnotesize (a) ECSIC} & \mbox{\footnotesize (b) JCT-only} & \mbox{\footnotesize (c) PE-only} & \mbox{\footnotesize (d) FCNR} & \mbox{\footnotesize (e) GT} 
  \end{array}$
 \end{center}
\vspace{-.25in} 
 \caption{First row: decompressed images of FCNR and ECSIC with variations. Second row: zoom-ins for closer examination.} 
 \label{fig:ablation-study}
\end{figure}
%\vspace{-.1in} 

Table~\ref{tab:ablation-study} shows that both modifications lead to improvements in image quality measured by PSNR (JCT-only and PE-only) and LPIPS (JCT-only). Moreover, though JCT-only leads to higher BPP than ECSIC, PE-only lowers BPP. By incorporating both modifications, FCNR achieves even lower BPP with the best PSNR and LPIPS.
Figure~\ref{fig:ablation-study} demonstrates visual improvements in image quality. JCT-only yields images with smoother surfaces and more natural lighting, and PE-only reduces visual artifacts to some extent and enhances lighting concentration. As the zoom-ins indicate, FCNR further improves border clearness, lighting concentration, and color consistency, generating an image closest to GT.

{\bf Discussion.}
Our results demonstrate that FCNR can compress a large collection of visualization images in high fidelity within a short time. It is much more promising than INR-based methods when image quality and encoding and decoding speed are of greater importance. Though ECSIC can achieve high-quality compression in a similar timeframe, FCNR leads to a higher compression ratio.

\vspace{-0.1in}
\section{Conclusions and Future Work}

We present FCNR, a novel method for neural compression of visualization images borrowing insights from stereo image compression frameworks. 
The model of ECSIC reduces the bitrate with distributions of the right image learned from the left image using SCMs. 
We integrate this model with JCTMs to extract mutual information globally and incorporate visualization parameters to allow for more detailed quantitative differences between images, further improving image quality and compression ratio. 
Compared with state-of-the-art INR-based methods, 
FCNR provides previously unavailable interpolation ability and demonstrates improved encoding and decoding time.
Compared with ECSIC, FCNR achieves a higher compression ratio and slightly better reconstruction quality. 
%Table~\ref{tab:comp-methods} summarizes the comparison of all methods.

The future work of FCNR can be summarized as follows. 
First, given the substantial differences between stereo images and visualization images, designing a more tailored model architecture for visualization images is necessary for further gains in quality and speed. 
Second, FCNR lags behind INR-based methods in terms of compression ratio. 
Our method will be more promising if BPP can be reduced to the same level as INR-based methods. 
Finally, more visualization parameters, such as isovalues and transfer functions, may be included, and a better fusion of these parameters with images is worthy of exploration.

\vspace{-0.1in}
\acknowledgments{This research was supported in part by the U.S.\ National Science Foundation through grants IIS-1955395, IIS-2101696, OAC-2104158, and IIS-2401144, and the U.S.\ Department of Energy through grant DE-SC0023145. The authors would like to thank the anonymous reviewers for their insightful comments.}

\vspace{-0.1in}
\bibliographystyle{abbrv-doi}

\bibliography{template}

\begin{thebibliography}{10}

\bibitem{balle2017end}
J.~Ball{\'e}, V.~Laparra, and E.~P. Simoncelli.
\newblock End-to-end optimized image compression.
\newblock In {\em Proceedings of International Conference on Learning Representations}, 2017.

\bibitem{balle2018variational}
J.~Ball{\'e}, D.~Minnen, S.~Singh, S.~J. Hwang, and N.~Johnston.
\newblock Variational image compression with a scale hyperprior.
\newblock In {\em Proceedings of International Conference on Learning Representations}, 2018.

\bibitem{chen2022cnerv}
H.~Chen, M.~Gwilliam, B.~He, S.-N. Lim, and A.~Shrivastava.
\newblock {CNeRV}: Content-adaptive neural representation for visual data.
\newblock In {\em Proceedings of British Machine Vision Conference}, pp. 510:1--510:20, 2022.

\bibitem{chen2023hnerv}
H.~Chen, M.~Gwilliam, S.-N. Lim, and A.~Shrivastava.
\newblock {HNeRV}: A hybrid neural representation for videos.
\newblock In {\em Proceedings of IEEE Conference on Computer Vision and Pattern Recognition}, pp. 10270--10279, 2023. doi: {{%
10\hspace{.1pt}\discretionary{.}{%
}{.}\hspace{.4pt}1109\discretionary{/}{%
}{/}CVPR52729\hspace{.1pt}\discretionary{.}{%
}{.}\hspace{.4pt}2023\hspace{.1pt}\discretionary{.}{%
}{.}\hspace{.4pt}00990}}


\bibitem{chen2021nerv}
H.~Chen, B.~He, H.~Wang, Y.~Ren, S.-N. Lim, and A.~Shrivastava.
\newblock {NeRV}: Neural representations for videos.
\newblock In {\em Proceedings of Advances in Neural Information Processing Systems}, pp. 21557--21568, 2021.

\bibitem{crawfis1993texture}
R.~A. Crawfis and N.~Max.
\newblock Texture splats for 3{D} scalar and vector field visualization.
\newblock In {\em Proceedings of IEEE Visualization Conference}, pp. 261--267, 1993. doi: {{%
10\hspace{.1pt}\discretionary{.}{%
}{.}\hspace{.4pt}1109\discretionary{/}{%
}{/}VISUAL\hspace{.1pt}\discretionary{.}{%
}{.}\hspace{.4pt}1993\hspace{.1pt}\discretionary{.}{%
}{.}\hspace{.4pt}398877}}


\bibitem{deng2021deep}
X.~Deng, W.~Yang, R.~Yang, M.~Xu, E.~Liu, Q.~Feng, and R.~Timofte.
\newblock Deep homography for efficient stereo image compression.
\newblock In {\em Proceedings of IEEE Conference on Computer Vision and Pattern Recognition}, pp. 1492--1501, 2021. doi: {{%
10\hspace{.1pt}\discretionary{.}{%
}{.}\hspace{.4pt}1109\discretionary{/}{%
}{/}CVPR46437\hspace{.1pt}\discretionary{.}{%
}{.}\hspace{.4pt}2021\hspace{.1pt}\discretionary{.}{%
}{.}\hspace{.4pt}00154}}


\bibitem{Gu-CG23}
P.~Gu, D.~Z. Chen, and C.~Wang.
\newblock {NeRVI}: Compressive neural representation of visualization images for communicating volume visualization results.
\newblock {\em Computers \& Graphics}, 116:216--227, 2023. doi: {{%
10\hspace{.1pt}\discretionary{.}{%
}{.}\hspace{.4pt}1016\discretionary{/}{%
}{/}J\hspace{.1pt}\discretionary{.}{%
}{.}\hspace{.4pt}CAG\hspace{.1pt}\discretionary{.}{%
}{.}\hspace{.4pt}2023\hspace{.1pt}\discretionary{.}{%
}{.}\hspace{.4pt}08\hspace{.1pt}\discretionary{.}{%
}{.}\hspace{.4pt}024}}


\bibitem{Han-TVCG23}
J.~Han and C.~Wang.
\newblock {CoordNet}: Data generation and visualization generation for time-varying volumes via a coordinate-based neural network.
\newblock {\em IEEE Transactions on Visualization and Computer Graphics}, 29(12):4951--4963, 2023. doi: {{%
10\hspace{.1pt}\discretionary{.}{%
}{.}\hspace{.4pt}1109\discretionary{/}{%
}{/}TVCG\hspace{.1pt}\discretionary{.}{%
}{.}\hspace{.4pt}2022\hspace{.1pt}\discretionary{.}{%
}{.}\hspace{.4pt}3197203}}


\bibitem{he2015prelu}
K.~He, X.~Zhang, S.~Ren, and J.~Sun.
\newblock Delving deep into rectifiers: Surpassing human-level performance on {ImageNet} classification.
\newblock In {\em Proceedings of IEEE International Conference on Computer Vision}, pp. 1026--1034, 2015. doi: {{%
10\hspace{.1pt}\discretionary{.}{%
}{.}\hspace{.4pt}1109\discretionary{/}{%
}{/}ICCV\hspace{.1pt}\discretionary{.}{%
}{.}\hspace{.4pt}2015\hspace{.1pt}\discretionary{.}{%
}{.}\hspace{.4pt}123}}


\bibitem{kwan2024hinerv}
H.~M. Kwan, G.~Gao, F.~Zhang, A.~Gower, and D.~Bull.
\newblock {HiNeRV}: Video compression with hierarchical encoding based neural representation.
\newblock In {\em Proceedings of Advances in Neural Information Processing Systems}, 2023.

\bibitem{li2022enerv}
Z.~Li, M.~Wang, H.~Pi, K.~Xu, J.~Mei, and Y.~Liu.
\newblock {E-NeRV}: Expedite neural video representation with disentangled spatial-temporal context.
\newblock In {\em Proceedings of European Conference on Computer Vision}, pp. 267--284, 2022. doi: {{%
10\hspace{.1pt}\discretionary{.}{%
}{.}\hspace{.4pt}1007\discretionary{/}{%
}{/}978\discretionary{%
}{-}{-}3\discretionary{%
}{-}{-}031\discretionary{%
}{-}{-}19833\discretionary{%
}{-}{-}5\_16}}


\bibitem{liu2019dsic}
J.~Liu, S.~Wang, and R.~Urtasun.
\newblock {DSIC}: Deep stereo image compression.
\newblock In {\em Proceedings of IEEE International Conference on Computer Vision}, pp. 3136--3145, 2019. doi: {{%
10\hspace{.1pt}\discretionary{.}{%
}{.}\hspace{.4pt}1109\discretionary{/}{%
}{/}ICCV\hspace{.1pt}\discretionary{.}{%
}{.}\hspace{.4pt}2019\hspace{.1pt}\discretionary{.}{%
}{.}\hspace{.4pt}00323}}


\bibitem{maiya2023nirvana}
S.~R. Maiya, S.~Girish, M.~Ehrlich, H.~Wang, K.~S. Lee, P.~Poirson, P.~Wu, C.~Wang, and A.~Shrivastava.
\newblock {NIRVANA}: Neural implicit representations of videos with adaptive networks and autoregressive patch-wise modeling.
\newblock In {\em Proceedings of IEEE Conference on Computer Vision and Pattern Recognition}, pp. 14378--14387, 2023. doi: {{%
10\hspace{.1pt}\discretionary{.}{%
}{.}\hspace{.4pt}1109\discretionary{/}{%
}{/}CVPR52729\hspace{.1pt}\discretionary{.}{%
}{.}\hspace{.4pt}2023\hspace{.1pt}\discretionary{.}{%
}{.}\hspace{.4pt}01382}}


\bibitem{minnen2020channel}
D.~Minnen and S.~Singh.
\newblock Channel-wise autoregressive entropy models for learned image compression.
\newblock In {\em Proceedings of IEEE International Conference on Image Processing}, pp. 3339--3343, 2020. doi: {{%
10\hspace{.1pt}\discretionary{.}{%
}{.}\hspace{.4pt}1109\discretionary{/}{%
}{/}ICIP40778\hspace{.1pt}\discretionary{.}{%
}{.}\hspace{.4pt}2020\hspace{.1pt}\discretionary{.}{%
}{.}\hspace{.4pt}9190935}}


\bibitem{popinet2004experimental}
S.~Popinet, M.~Smith, and C.~Stevens.
\newblock Experimental and numerical study of the turbulence characteristics of airflow around a research vessel.
\newblock {\em Journal of Atmospheric and Oceanic Technology}, 21(10):1575--1589, 2004. doi: {{%
10\hspace{.1pt}\discretionary{.}{%
}{.}\hspace{.4pt}1175\discretionary{/}{%
}{/}1520\discretionary{%
}{-}{-}0426\discretionary{%
}{(}{(}2004\discretionary{)}{%
}{)}021{\textless}1575\discretionary{:}{%
}{:}EANSOT{\textgreater}2\hspace{.1pt}\discretionary{.}{%
}{.}\hspace{.4pt}0\hspace{.1pt}\discretionary{.}{%
}{.}\hspace{.4pt}CO\discretionary{;}{%
}{;}2}}


\bibitem{silver1997tracking}
D.~Silver and X.~Wang.
\newblock Tracking and visualizing turbulent {3D} features.
\newblock {\em IEEE Transactions on Visualization and Computer Graphics}, 3(2):129--141, 1997. doi: {{%
10\hspace{.1pt}\discretionary{.}{%
}{.}\hspace{.4pt}1109\discretionary{/}{%
}{/}2945\hspace{.1pt}\discretionary{.}{%
}{.}\hspace{.4pt}597796}}


\bibitem{Tang-PVIS24}
K.~Tang and C.~Wang.
\newblock {ECNR}: Efficient compressive neural representation of time-varying volumetric datasets.
\newblock In {\em Proceedings of IEEE Pacific Visualization Conference}, pp. 72--81, 2024. doi: {{%
10\hspace{.1pt}\discretionary{.}{%
}{.}\hspace{.4pt}1109\discretionary{/}{%
}{/}PACIFICVIS60374\hspace{.1pt}\discretionary{.}{%
}{.}\hspace{.4pt}2024\hspace{.1pt}\discretionary{.}{%
}{.}\hspace{.4pt}00017}}


\bibitem{Tang-CG24}
K.~Tang and C.~Wang.
\newblock {STSR-INR}: Spatiotemporal super-resolution for time-varying multivariate volumetric data via implicit neural representation.
\newblock {\em Computers \& Graphics}, 119:103874, 2024. doi: {{%
10\hspace{.1pt}\discretionary{.}{%
}{.}\hspace{.4pt}1016\discretionary{/}{%
}{/}J\hspace{.1pt}\discretionary{.}{%
}{.}\hspace{.4pt}CAG\hspace{.1pt}\discretionary{.}{%
}{.}\hspace{.4pt}2024\hspace{.1pt}\discretionary{.}{%
}{.}\hspace{.4pt}01\hspace{.1pt}\discretionary{.}{%
}{.}\hspace{.4pt}001}}


\bibitem{Wang-TVCG23}
C.~Wang and J.~Han.
\newblock {DL4SciVis}: A state-of-the-art survey on deep learning for scientific visualization.
\newblock {\em IEEE Transactions on Visualization and Computer Graphics}, 29(8):3714--3733, 2023. doi: {{%
10\hspace{.1pt}\discretionary{.}{%
}{.}\hspace{.4pt}1109\discretionary{/}{%
}{/}TVCG\hspace{.1pt}\discretionary{.}{%
}{.}\hspace{.4pt}2022\hspace{.1pt}\discretionary{.}{%
}{.}\hspace{.4pt}3167896}}


\bibitem{wodlinger2024ecsic}
M.~W{\"o}dlinger, J.~Kotera, M.~Keglevic, J.~Xu, and R.~Sablatnig.
\newblock {ECSIC}: Epipolar cross attention for stereo image compression.
\newblock In {\em Proceedings of IEEE Winter Conference on Applications of Computer Vision}, pp. 3436--3445, 2024.

\bibitem{wodlinger2022sasic}
M.~W{\"o}dlinger, J.~Kotera, J.~Xu, and R.~Sablatnig.
\newblock {SASIC}: Stereo image compression with latent shifts and stereo attention.
\newblock In {\em Proceedings of IEEE Conference on Computer Vision and Pattern Recognition}, pp. 661--670, 2022. doi: {{%
10\hspace{.1pt}\discretionary{.}{%
}{.}\hspace{.4pt}1109\discretionary{/}{%
}{/}CVPR52688\hspace{.1pt}\discretionary{.}{%
}{.}\hspace{.4pt}2022\hspace{.1pt}\discretionary{.}{%
}{.}\hspace{.4pt}00074}}


\bibitem{Zhang-CVPR18}
R.~Zhang, P.~Isola, A.~A. Efros, E.~Shechtman, and O.~Wang.
\newblock The unreasonable effectiveness of deep features as a perceptual metric.
\newblock In {\em Proceedings of IEEE Conference on Computer Vision and Pattern Recognition}, pp. 586--595, 2018. doi: {{%
10\hspace{.1pt}\discretionary{.}{%
}{.}\hspace{.4pt}1109\discretionary{/}{%
}{/}CVPR\hspace{.1pt}\discretionary{.}{%
}{.}\hspace{.4pt}2018\hspace{.1pt}\discretionary{.}{%
}{.}\hspace{.4pt}00068}}


\bibitem{zhang2023ldmic}
X.~Zhang, J.~Shao, and J.~Zhang.
\newblock {LDMIC}: Learning-based distributed multi-view image coding.
\newblock In {\em Proceedings of International Conference on Learning Representations}, 2023.

\bibitem{zhao2023dnerv}
Q.~Zhao, M.~S. Asif, and Z.~Ma.
\newblock {DNeRV}: Modeling inherent dynamics via difference neural representation for videos.
\newblock In {\em Proceedings of IEEE Conference on Computer Vision and Pattern Recognition}, pp. 2031--2040, 2023. doi: {{%
10\hspace{.1pt}\discretionary{.}{%
}{.}\hspace{.4pt}1109\discretionary{/}{%
}{/}CVPR52729\hspace{.1pt}\discretionary{.}{%
}{.}\hspace{.4pt}2023\hspace{.1pt}\discretionary{.}{%
}{.}\hspace{.4pt}00202}}


\end{thebibliography}
\end{document}